\pdfoutput=1

\documentclass{article}

\usepackage[utf8]{inputenc}
\usepackage[a4paper, total={6.8in, 9.5in}]{geometry}

\usepackage[hyphens]{url}
\usepackage{hyperref}

\usepackage{tikzsymbols}
\usetikzlibrary{backgrounds}

\usepackage{authblk}
\usepackage{tikz}
\usepackage{multirow}
\usepackage{amsmath}
\usepackage{amsfonts}

\usepackage{media9}
\definecolor{darkgreen}{rgb}{0.0, 0.2, 0.13}

\usepackage{listings}
\usepackage{graphicx} 
\usepackage{pgfplots}
\pgfplotsset{compat=1.14}

\usepackage{subcaption}

\definecolor{ao}{rgb}{0.0, 0.5, 0.0}

\usetikzlibrary{positioning}
\usetikzlibrary{calc}

\usepackage[title]{appendix}

\usepackage{mathtools}

\tikzset{mybox/.style={draw, minimum width=1cm, fill=gray!10}}

\tikzset{myboxLeading/.style={draw, minimum width=0.7cm, fill=yellow!10}}

\tikzset{myboxA/.style={draw, minimum width=0.7cm, fill=blue!10}}

\tikzset{myboxFst/.style={draw, minimum width=0.5cm, fill=yellow!10}}
\tikzset{myboxSec/.style={draw, minimum width=0.5cm, fill=green!10}}
\tikzset{myboxLst/.style={draw, minimum width=0.5cm, fill=pink!10}}
\tikzset{myboxLstFinal/.style={draw, minimum width=0.5cm, fill=blue!10}}

\tikzset{boxBasis/.style={draw, minimum width=1cm, fill=red!10}}

\usetikzlibrary{shapes}

\usetikzlibrary{decorations.shapes}
\tikzset{decorate sep/.style 2 args=
{decorate,decoration={shape backgrounds,shape=circle,shape size=#1,shape sep=#2}}}

\usepackage{xcolor}

\newcommand{\highlight}[2][yellow]{\mathchoice%
  {\colorbox{#1}{$\displaystyle#2$}}%
  {\colorbox{#1}{$\textstyle#2$}}%
  {\colorbox{#1}{$\scriptstyle#2$}}%
  {\colorbox{#1}{$\scriptscriptstyle#2$}}}%

\makeatletter
\DeclareRobustCommand*{\Myhref}[1][]{
  \begingroup
  \setkeys{Hyp}{#1}
  \@ifnextchar\bgroup\Hy@href{\hyper@normalise\href@}%
}
\makeatother

\definecolor{codegreen}{rgb}{0,0.6,0}
\definecolor{codegray}{rgb}{0.5,0.5,0.5}
\definecolor{codepurple}{rgb}{0.58,0,0.82}
\definecolor{backcolour}{rgb}{0.95,0.95,0.92}

\lstdefinestyle{mystyle}{
  backgroundcolor=\color{backcolour},   commentstyle=\color{codegreen},
  keywordstyle=\color{magenta},
  numberstyle=\tiny\color{codegray},
  stringstyle=\color{codepurple},
  basicstyle=\footnotesize,
  breakatwhitespace=false,         
  breaklines=true,                 
  captionpos=b,                    
  keepspaces=true,                 
  numbers=left,                    
  numbersep=5pt,                  
  showspaces=false,                
  showstringspaces=false,
  showtabs=false,                  
  tabsize=2,
  linewidth=17cm
}

\begin{document}

\title{Real numbers, data science and chaos: \\  How to fit any dataset with a single parameter}
\author{Laurent Bou\'e \\ \Myhref[hidelinks]{mailto:ranlot75@gmail.com}{\protect\includegraphics[width=1cm,height=0.6cm]{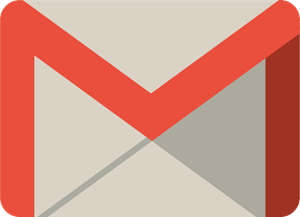}} \,\,\,\, \Myhref[hidelinks]{https://www.linkedin.com/in/laurent-bou\%C3\%A9-b7923853/}{\protect\includegraphics[width=0.8cm,height=0.8cm]{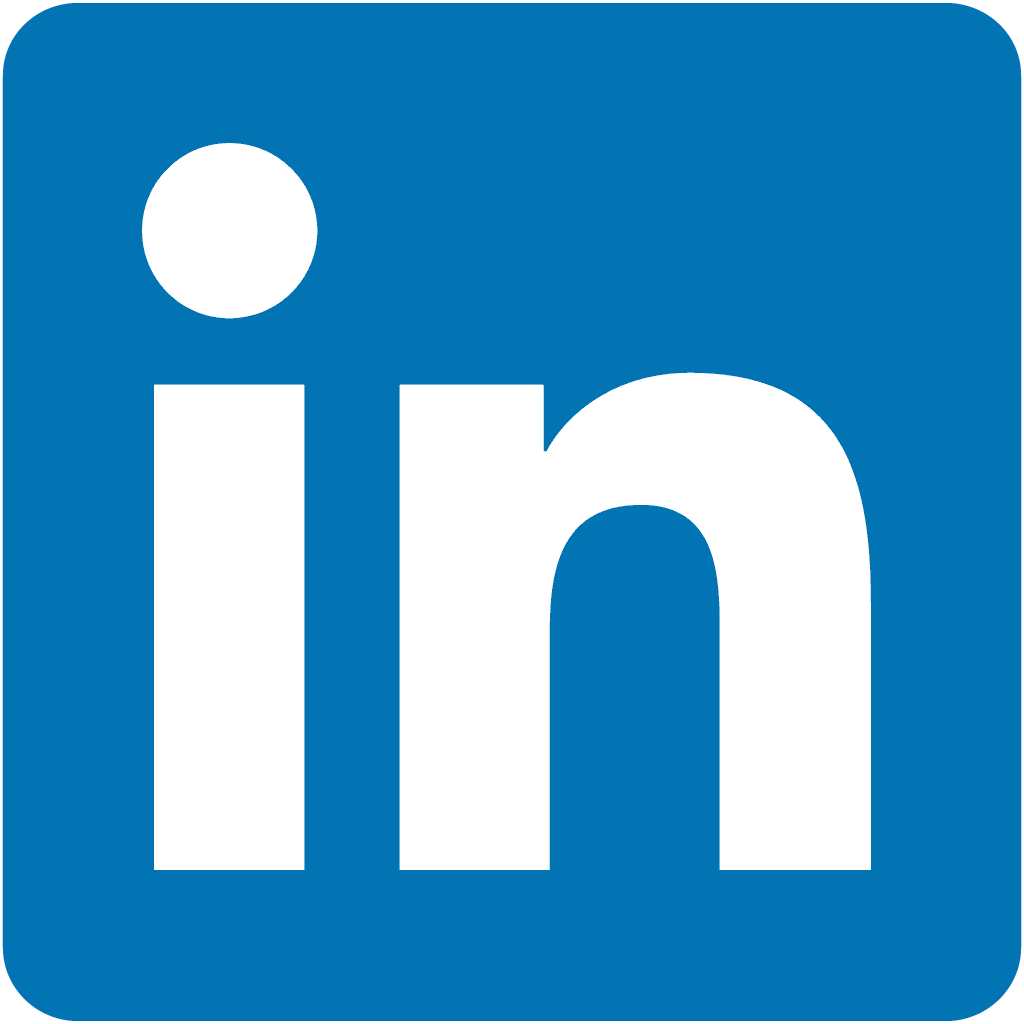}} \,\,\,\, \Myhref[hidelinks]{https://github.com/Ranlot}{\protect\includegraphics[width=1cm,height=1cm]{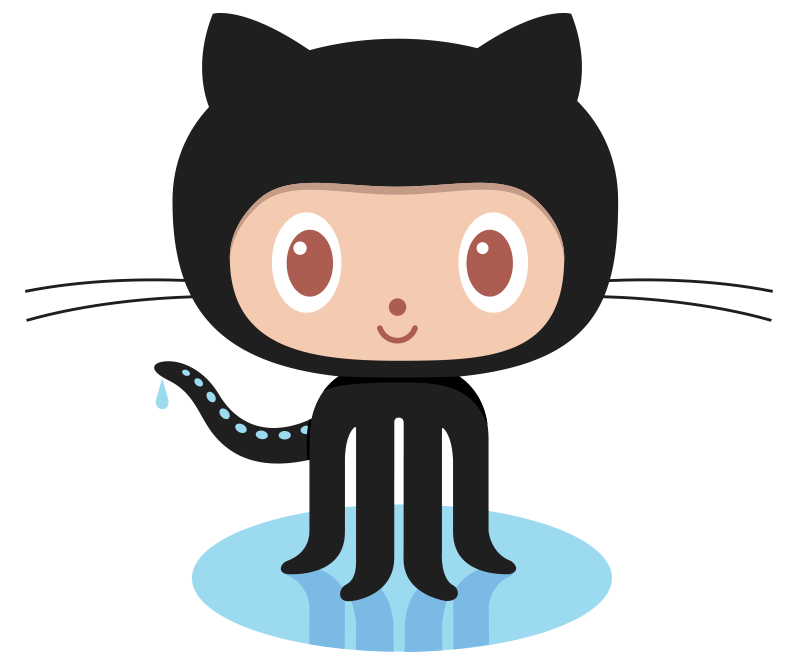}} \,\,\,\, \Myhref[hidelinks]{https://twitter.com/ranlot75}{\protect\includegraphics[width=0.8cm,height=0.8cm]{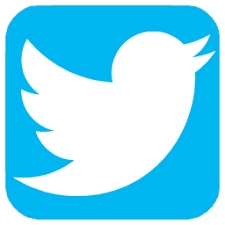}} }
\affil{SAP Labs}
\date{}
\maketitle

\begin{abstract}
We show how any dataset of any modality (time-series, images, sound...) can be approximated by a well-behaved (continuous, differentiable...) scalar function with a single real-valued parameter.  Building upon elementary concepts from chaos theory, we adopt a pedagogical approach demonstrating how to adjust this parameter in order to achieve arbitrary precision fit to all samples of the data.   Targeting an audience of data scientists with a taste for the curious and unusual, the results presented here expand on previous similar observations~\cite{piantadosi} regarding expressiveness power and generalization of machine learning models.
\end{abstract}

\vskip 0.2in


\vskip 0.2in

\begin{figure}[hb!]
\centering
\includegraphics[width=.245\linewidth]{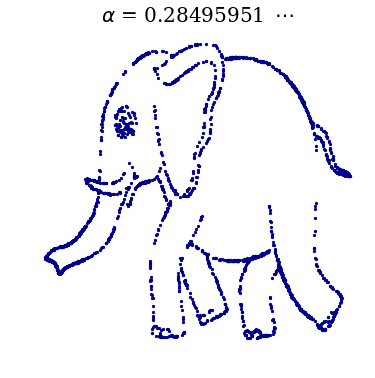}
\includegraphics[width=.245\linewidth]{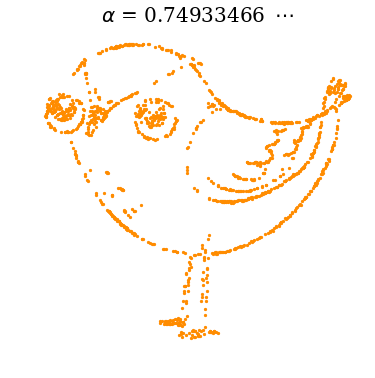}
\includegraphics[width=.245\linewidth]{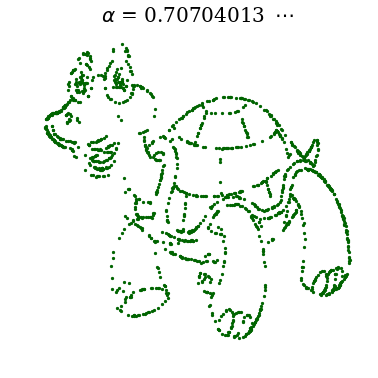}
\includegraphics[width=.245\linewidth]{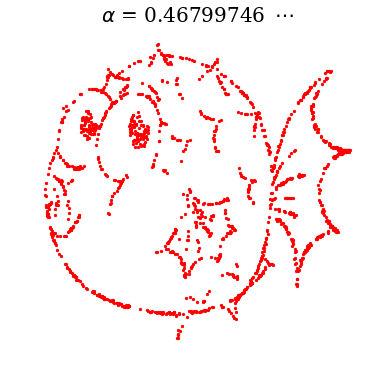}
\caption{Animal shapes obtained with the different values of~$\alpha$ defined on top of each image.  One should consider the data as a scatter plot of pairs of values~$(x,y)$ where each~$x \in \mathbb{N}$ is associated with a corresponding~$y$ value given by~$y \equiv f_\alpha(x)$.  One goal of the paper will be to show how to find the precise value of~$\alpha \in \mathbb{R}$ required to fit any target dataset.}
\label{fig::animals}
\end{figure}

Real world data comes in huge variety of shapes and sizes with modalities ranging from traditional structured database schemas to unstructured media sources such as video feeds and audio recordings.  Nevertheless, any dataset can ultimately be thought of as a list of numerical values~$\mathcal{X} = [x_0, \cdots , x_n]$ describing the data content regardless of the underlying modality.  The purpose of this paper is to show that all the samples of any arbitrary dataset~$\mathcal{X}$ can be reproduced by a simple differentiable equation:
\begin{equation}
f_\alpha(x) = \sin^2 \left( 2^{x\tau} \arcsin{\sqrt{\alpha}} \right)
\label{eq::mainResult}
\end{equation}
where~$\alpha \in \mathbb{R}$ is a real-valued parameter to be learned from the data and~$x \in [0, \cdots , n]$ takes integer values.  ($\tau \in \mathbb{N}$ is a constant which effectively controls the desired level of accuracy).  Before delving into the logic of how and why~$f_\alpha$ is able to achieve such a lofty goal, let us start by a few practical demonstrations.  Keeping with the tradition of ``fitting an elephant''~\cite{elephant}, we start by showing how different animal shapes may be generated by choosing appropriate values of~$\alpha$ as displayed in~Figure~\ref{fig::animals}. \\

\begin{figure}
\centering
\begin{subfigure}{.5\textwidth}
  \centering
  \includegraphics[width=\linewidth]{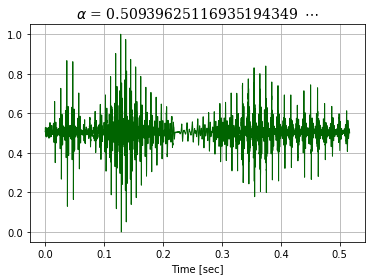}
\end{subfigure}%
\begin{subfigure}{.5\textwidth}
  \centering
  \includegraphics[width=\linewidth]{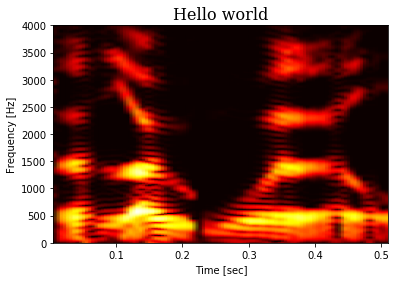}
\end{subfigure}
\caption{{\bf Left)} Time series obtained by applying~$f_\alpha$ at equally-spaced arguments in~$\mathbb{N}$.  Scaling the horizontal axis by an appropriate sampling rate ($\approx 11000$~Hz) allows the signal to be interpreted as a sound wave. {\bf Right)}~Playing the corresponding \includemedia[
  transparent,
  passcontext,
  addresource=resources/audio/generated/HelloWorld_generated.mp3,
  flashvars={source=resources/audio/generated/HelloWorld_generated.mp3},
]{\color{darkgreen}\framebox[0.1\linewidth][c]{\bf audio file}}{APlayer.swf} reveals a male voice saying~``Hello world'' characterized by this spectrogram.}
\label{fig:helloWorld}
\end{figure}

\noindent Following this demonstration that~$f_\alpha$ can generate any kind of doodle-like drawing, let us continue with a literal ``Hello world'' example in order to further illustrate the capabilities of the approach. Namely, we show in~Figure~\ref{fig:helloWorld} how a well-chosen value of~$\alpha$ may be used to produce a complex high-dimensional acoustic signal encoding the actual expression~``Hello world''! \\

\begin{figure}
\centering
\includegraphics[width=\linewidth]{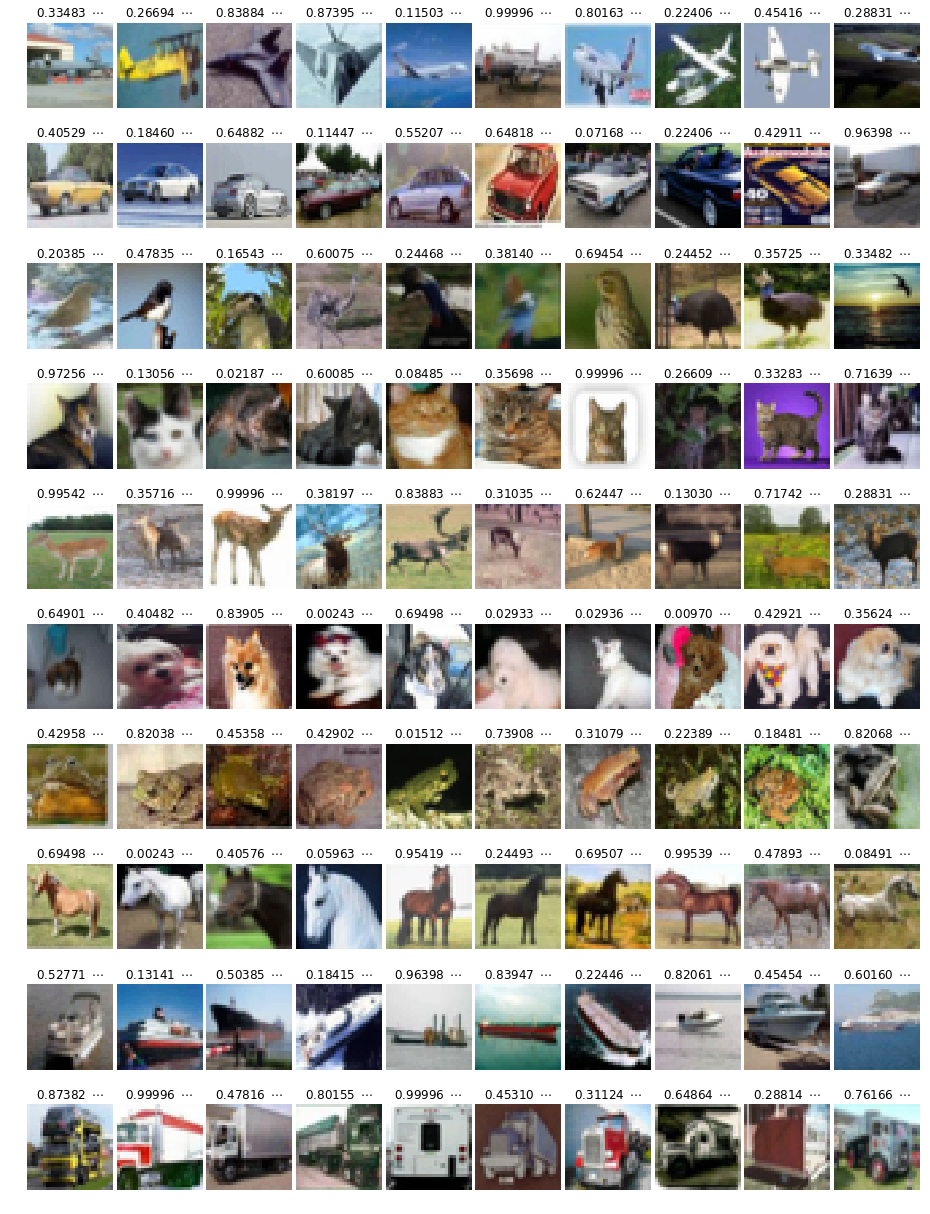}
\caption[]{Small resolution images are generated by applying~$f_\alpha$ for~$3072$ integer-spaced arguments in~$\mathbb{N}$.  Folding the resulting list of numerical values appropriately into~$3$d arrays of shapes~$32\times 32\times 3$ allows one to produce color images that look like they were drawn from the~CIFAR-10 classes: \{{\it airplane, automobile, bird, cat, deer, dog, frog, horse, ship, truck}\}.}
\label{fig::cifar10}
\end{figure}

\noindent Moving on to the connection with data science, let us consider the data modality that has emerged as the epicenter of the current explosion of interest in deep learning: images.  Along with the advent of specialized hardware and smarter neural network architectures, it is widely acknowledged that the availability of very large labeled training data has been one of the most important factor responsible for the perceived computer vision ``coming-of-age'' as manifested by a myriad of academic breakthroughs and integration into successful commercial applications.  In this context, the CIFAR-10 dataset continues to stand as one influential yardstick measuring the performance of new learning algorithms.  Because of the small resolution of the images, this dataset also serves as a popular research playground.  Accordingly, we demonstrate in~Figure~\ref{fig::cifar10} that it is always possible to find values of~$\alpha$ such that~$f_\alpha$ builds up artificial images that mirror the categories of~CIFAR-10.

\paragraph{Model complexity and generalization?} A fundamental question in machine learning research has to do with the evaluation of ``model capacity'' and its connection to the presumed existence of generalization (even in strongly overparametrized models). Traditional approaches tend to fall in categories such as \href{https://en.wikipedia.org/wiki/Vapnik-Chervonenkis\_dimension}{VC dimension} and \href{https://en.wikipedia.org/wiki/Rademacher\_complexity}{Rademacher complexity}...  Nevertheless, in the absence of a generic theoretical framework valid for modern deep learning architectures, it is not uncommon for practitioners to simply count the number of parameters and treat it as a proxy for expressiveness power; an approach somehow inspired by classical information theory based estimators such as AIC and BIC. \\

\noindent The examples above have demonstrated that an elegant model~$f_\alpha$ with a simple and differentiable formulation (composition of a few trigonometric and exponential functions) is able to produce any kind of semantically-relevant scatter plot, audio or visual data (text may also be constructed using an identical approach) at the cost of a single real-valued parameter.  Without yet revealing all the tricks, one aspect should already be obvious by now: all the information is directly encoded, without any compression or ``learning'', into~$\alpha \in \mathbb{R}$.  As mathematical objects, real numbers are non-terminating and therefore contain an infinite amount of information (in particular, they should not be confused with whatever finite-precision data types programming languages may implement)~\cite{chaitin}.  As such, no generalization can be expected from~$f_\alpha$ and indeed we conclude the paper by showing this explicitly in the context of time-series in~Figure~\ref{fig:generalization}. \\

\noindent In addition to casting doubts on the validity of parameter-counting methods and highlighting the importance of complexity bounds based on Occam's razor such as~\href{https://en.wikipedia.org/wiki/Minimum\_description\_length}{minimum description length} (that trade off goodness-of-fit with expressive power), we hope that~$f_\alpha$ may also serve as {\bf entertainment for curious data scientists}~\cite{laurent}. 

\paragraph{Setting up a few essential building blocks.}  After revealing that~$\alpha$ directly encodes the entire dataset~$\mathcal{X}$ into a single parameter, the rest of the paper will be dedicated to showing how to construct~$\alpha \in \mathbb{R}$ and its companion decoder function~$f_\alpha$.  In order to do this, we need to lay out a couple of conceptual tools upon which the whole trick is based.

\paragraph{\framebox{\it a) Fixed-point binary representation}}  Without loss of generality, we focus on real numbers~$\alpha \in [0,1]$ in the unit interval.  Since the integral part is identically null, the fractional part of any such~$\alpha$ can be expressed as an infinite vector of coefficients~$a_i = \{0,1\}$ where each coefficient is paired with a weight~$1/2^i$ as illustrated by:

\begin{center} \begin{tikzpicture}
\node[mybox] (a1) at (9, 5.45) {$1/2^1$};  
\node[mybox] (a2) at (10, 5.45) {$1/2^2$};
\node[mybox] (a3) at (11, 5.45) {$1/2^3$};
\node[mybox] (a4) at (12, 5.45) {$1/2^4$};
\node[mybox] (a5) at (13, 5.45) {$1/2^5$};
\node[mybox] (a6) at (14, 5.45) {$1/2^6$};
\node[mybox] (a7) at (15, 5.45) {$1/2^7$};
\node[mybox] (a8) at (16, 5.45) {$1/2^8$};
\node[mybox] (a9) at (17, 5.45) {$1/2^9$};
\node[mybox] (a10) at (18, 5.45) {$1/2^{10}$};
\node[mybox] (a11) at (19, 5.45) {$1/2^{11}$};
\node[fill=gray!10] (dots1) at (19.9, 5.45) {$\cdots$};
\node[fill=gray!10] (dots2) at (20.4, 5.45) {$\cdots$};
\node[fill=gray!10] (dots3) at (20.9, 5.45) {$\cdots$};

\node[myboxLeading] (ex0) at (7.7, 4.6) {$\alpha = 0 \, . $};
\node[boxBasis] (exs1) at (9, 4.6) {$a_1$};  
\node[boxBasis] (exs2) at (10, 4.6) {$a_2$};
\node[boxBasis] (exs3) at (11, 4.6) {$a_3$};
\node[boxBasis] (exs4) at (12, 4.6) {$a_4$};
\node[boxBasis] (exs5) at (13, 4.6) {$a_5$};
\node[boxBasis] (exs6) at (14, 4.6) {$a_6$};
\node[boxBasis] (exs7) at (15, 4.6) {$a_7$};
\node[boxBasis] (exs8) at (16, 4.6) {$a_8$};
\node[boxBasis] (exs9) at (17, 4.6) {$a_9$};
\node[boxBasis] (exs10) at (18, 4.6) {$a_{10}$};
\node[boxBasis] (exs11) at (19, 4.6) {$a_{11}$};
\node[fill=red!10] (adots1) at (19.9, 4.6) {$\cdots$};
\node[fill=red!10] (adots2) at (20.4, 4.6) {$\cdots$};
\node[fill=red!10] (adots3) at (20.9, 4.6) {$\cdots$};

\end{tikzpicture} \end{center}
Converting this binary representation of~$\alpha$ to its equivalent decimal counterpart is accomplished by evaluating the following infinite-sum expression:
\begin{equation*}
\alpha = \sum_{n=1}^{+\infty} \dfrac{a_n}{2^n} \,\, ; \quad \,\, \alpha \in [0,1]
\end{equation*}
Unfortunately, existing data warehouse solutions can only store a finite (however large) amount of information and cannot handle the infinite memory requirements imposed by mathematical real numbers in~$\mathbb{R}$.  Instead, one has to decide on a threshold to approximate real numbers thereby resulting in a finite~$\tau$-bit expansion: \\

\noindent \makebox[\textwidth][c]{\begin{tikzpicture}

\node[myboxLeading] () at (8, 3) {$0 \, . $};
\node[boxBasis] (a1) at (9, 3) {$0$};  
\node[boxBasis] () at (10, 3) {$0$};
\node[boxBasis] () at (11, 3) {$0$};
\node[boxBasis] () at (12, 3) {$0$};
\node[boxBasis] () at (13, 3) {$0$};
\node[boxBasis] () at (14, 3) {$0$};
\node[boxBasis] () at (15, 3) {$0$};
\node[boxBasis] (a8) at (16, 3) {$0$};
\node[text width=2.15cm] (xx) at (17.8, 3) {$ = 0$};

\node[myboxLeading] () at (8, 2.4) {$0 \, . $};
\node[boxBasis] () at (9, 2.4) {$0$};  
\node[boxBasis] () at (10, 2.4) {$0$};
\node[boxBasis] () at (11, 2.4) {$0$};
\node[boxBasis] () at (12, 2.4) {$0$};
\node[boxBasis] () at (13, 2.4) {$0$};
\node[boxBasis] () at (14, 2.4) {$0$};
\node[boxBasis] () at (15, 2.4) {$0$};
\node[boxBasis] () at (16, 2.4) {$1$};
\node[text width=2.15cm] () at (17.8, 2.4) {$ = 0.00390625$};

\node[] (dots) at (12.5, 1.8) {$\vdots$};

\node[myboxLeading] () at (8, 1.1) {$0 \, . $};
\node[boxBasis] () at (9, 1.1) {$1$};  
\node[boxBasis] () at (10, 1.1) {$0$};
\node[boxBasis] () at (11, 1.1) {$0$};
\node[boxBasis] () at (12, 1.1) {$0$};
\node[boxBasis] () at (13, 1.1) {$0$};
\node[boxBasis] () at (14, 1.1) {$0$};
\node[boxBasis] () at (15, 1.1) {$0$};
\node[boxBasis] () at (16, 1.1) {$0$};
\node[text width=2.15cm] () at (17.8, 1.1) {$ = 0.5$};

\node[] (dots) at (12.5, 0.5) {$\vdots$};

\node[myboxLeading] () at (8, -0.2) {$0 \, . $};
\node[boxBasis] () at (9, -0.2) {$1$};  
\node[boxBasis] () at (10, -0.2) {$1$};
\node[boxBasis] () at (11, -0.2) {$1$};
\node[boxBasis] () at (12, -0.2) {$1$};
\node[boxBasis] () at (13, -0.2) {$1$};
\node[boxBasis] () at (14, -0.2) {$1$};
\node[boxBasis] () at (15, -0.2) {$1$};
\node[boxBasis] () at (16, -0.2) {$0$};
\node[text width=2.15cm] () at (17.8, -0.2) {$ = 0.9921875$};

\node[myboxLeading] () at (8, -0.8) {$0 \, . $};
\node[boxBasis] () at (9, -0.8) {$1$};  
\node[boxBasis] () at (10, -0.8) {$1$};
\node[boxBasis] () at (11, -0.8) {$1$};
\node[boxBasis] () at (12, -0.8) {$1$};
\node[boxBasis] () at (13, -0.8) {$1$};
\node[boxBasis] () at (14, -0.8) {$1$};
\node[boxBasis] () at (15, -0.8) {$1$};
\node[boxBasis] (y) at (16, -0.8) {$1$};
\node[text width=2.15cm] (yy) at (17.8, -0.8) {$ = 0.9960938$};

\draw [decorate,decoration={brace,amplitude=10pt}] (a1.north west) -- (a8.north east) node [black,midway,yshift=0.3cm, above] { Truncation to finite~$\tau$-bit precision ($\tau = 8$)};

\draw [align=left, decorate,decoration={brace,amplitude=10pt}] (xx.north east) -- (yy.south east) node [black,midway,xshift=0.6cm, right] { Discretization of the unit interval \\  into $2^{\tau=8} = 256$ uniformly spaced \\  representable numbers. \\ \\  Empirical data points are rounded \\  off to the nearest available number \\  in this fixed-point representation.};

\end{tikzpicture}} \\

\noindent The equivalent decimal value is obtained by evaluating the now-finite sum:
\begin{equation*}
\alpha_\text{approx} = \sum_{n=1}^\tau \dfrac{a_n}{2^n}
\end{equation*}
This approximation is completely faithful only in the degenerate case where all components~$a_{n \geq \tau + 1} \equiv 0$. As soon as one of these components is non-zero~$\alpha_{\text{approx}}$ always underestimates the true value~$\alpha$.  In the worst case scenario, all the neglected components are~$a_{n \geq \tau + 1} \equiv 1$ showing that the error is bounded by:
\begin{equation*}
\left| \alpha - \alpha_\text{approx} \right| \leq \sum_{n=\tau+1}^{+\infty} \dfrac{1}{2^n} = \dfrac{1}{2^\tau}
\end{equation*}

\noindent This type of uniform discretization of the reals is referred to as a ``fixed-point'' representation to be contrasted with ``floating-point'' numbers that can represent both large and small reals in a reasonable amount of storage using a system similar to scientific notation.  Let us mention, in passing, that with the surge of compute-intensive workloads based on neural networks, research towards reducing hardware overhead has also sparked a renewed wave of interest in alternative number representations such as the ``logarithmic number system''~\cite{lns}...

\newpage

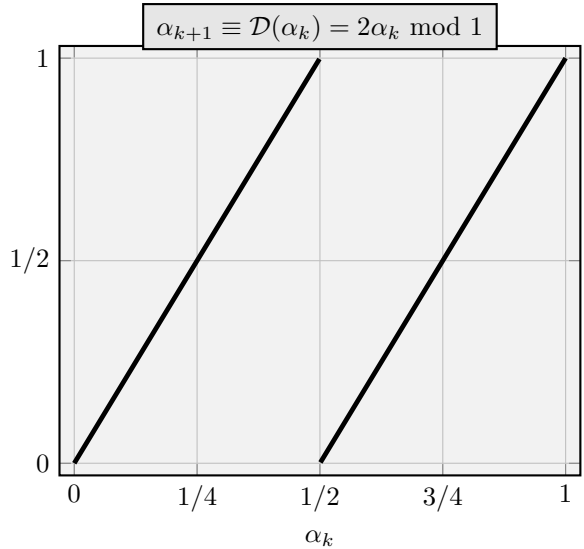
\begin{figure}
\begin{center}
\begin{tikzpicture} 
\begin{axis}[xmin=-0.03, xmax=1.03, ymin=-0.03, ymax=1.03, legend pos=north west, legend style={empty legend, fill=black!10, at={(0.5,1.1)},anchor=north}, axis background/.style={fill=gray!10}, xtick={0, 0.25, 0.5, 0.75, 1}, xticklabels={0, 1/4, 1/2, 3/4, 1},
ytick={0, 0.5, 1}, yticklabels={0, 1/2, 1}, thick, minor x tick num=0, minor y tick num=0, grid=both, xlabel=$\alpha_k$]
\addplot[samples=500,domain=0:0.499, color=black, line width=1.8pt]{2 * x};
\addplot[samples=500,domain=0.501:1, color=black, line width=1.8pt]{2 * x - 1};
\addlegendentry{$\alpha_{k+1} \equiv \mathcal{D} (\alpha_k) = 2 \alpha_k \,\,\text{mod}\,\,1$};
\end{axis}
\end{tikzpicture}
\end{center} 
\caption{Graphical illustration of the dyadic transformation~$\mathcal{D}$.  Notice how the modulo operation ensures that~$\alpha_{k+1} \in [0,1] \,\,\,\, \forall k \in \mathbb{N}$ by creating a non-differentiable jump at~$1/2$ turning~$\mathcal{D}$ into a piecewise linear function.}
\label{fig:dyadic}
\end{figure}

\paragraph{\framebox{\it b) Chaos theory}} The second (and last) prerequisite consists of a particularly simple incarnation of a one-dimensional discrete dynamical system known as the ``dyadic transformation''.  Given a variable~$\alpha_k \in [0,1]$ at time-step~$k$, its evolution at time-step~$k+1$ is defined by:
\begin{equation}
\alpha_{k+1} \equiv \mathcal{D} (\alpha_k) = 2 \alpha_k \,\, \text{mod} \,\, 1
\label{eq:dyadic}
\end{equation}

\noindent Stepping away from decimal representation and moving over to binary form helps reveal the underlying behavior of~$\mathcal{D}$ (illustrated graphically in Figure~\ref{fig:dyadic}).  Indeed, in this representation, multiplication by~2 can be interpreted as a simple shift of the whole bit sequence defining~$\alpha_k$ by a single bit in the leftwise direction.  In addition, the modulo operation ensures that every bit coming over on to the left-side of the radix point gets turned into a~$0$ thereby ensuring that~$\alpha_{k+1}$ remains in the unit interval.  Because of this property, the dyadic transformation is sometimes referred to as the ``bit-shift'' map.  As an example, let us consider~$\tau=8$ significant bits and denote by~$\highlight[red!10]{\cdots \cdots \cdots}$ the infinite sequence of random bits that follow.  Iterated applications of~$\mathcal{D}$ can be understood as an accumulation of leftwise bit-shifts:

\begin{center} \begin{tikzpicture}

\draw [dashed] (8.4, 3.5) -- (8.4, -2.7);

\node[myboxA] (ex0) at (3.6, 3.3) {$\alpha_k$};
\node[myboxLeading] (ex0) at (7.95, 3.3) {$0.$};
\node[boxBasis] (a1) at (9, 3.3) {$a_1$};  
\node[boxBasis] (a2) at (10, 3.3) {$a_2$};
\node[boxBasis] (a3) at (11, 3.3) {$a_3$};
\node[boxBasis] (a4) at (12, 3.3) {$a_4$};
\node[boxBasis] (a5) at (13, 3.3) {$a_5$};
\node[boxBasis] (a6) at (14, 3.3) {$a_6$};
\node[boxBasis] (a7) at (15, 3.3) {$a_7$};
\node[boxBasis] (a8) at (16, 3.3) {$a_8$};
\node[fill=red!10] () at (16.9, 3.3) {$\cdots$};
\node[fill=red!10] () at (17.4, 3.3) {$\cdots$};
\node[fill=red!10] () at (17.9, 3.3) {$\cdots$};

\node[myboxA] (ex0) at (4.5, 2.4) {$\alpha_{k+1} = \mathcal{D}^1(\alpha_k)$};
\node[myboxLeading] (ex0) at (7.15, 2.4) {$a_1\, \text{mod}\, 1 \equiv 0.$};
\node[boxBasis] (b2) at (9, 2.4) {$a_2$};  
\node[boxBasis] (b3) at (10, 2.4) {$a_3$};
\node[boxBasis] (b4) at (11, 2.4) {$a_4$};
\node[boxBasis] (b5) at (12, 2.4) {$a_5$};
\node[boxBasis] (b6) at (13, 2.4) {$a_6$};  
\node[boxBasis] (b7) at (14, 2.4) {$a_7$};
\node[boxBasis] (b8) at (15, 2.4) {$a_8$};
\node[fill=red!10] () at (15.9, 2.4) {$\cdots$};
\node[fill=red!10] () at (16.4, 2.4) {$\cdots$};
\node[fill=red!10] () at (16.9, 2.4) {$\cdots$};

\draw[->] (a2) to (b2) node[ ] {};
\draw[->] (a3) to (b3) node[ ] {};
\draw[->] (a4) to (b4) node[ ] {};
\draw[->] (a5) to (b5) node[ ] {};
\draw[->] (a6) to (b6) node[ ] {};
\draw[->] (a7) to (b7) node[ ] {};
\draw[->] (a8) to (b8) node[ ] {};

\node[myboxA] (ex0) at (4.5, 1.5) {$\alpha_{k+2} = \mathcal{D}^2(\alpha_k)$};
\node[myboxLeading] (ex0) at (7.15, 1.5) {$a_2\, \text{mod}\, 1 \equiv 0.$};
\node[boxBasis] (c3) at (9, 1.5) {$a_3$};  
\node[boxBasis] (c4) at (10, 1.5) {$a_4$};
\node[boxBasis] (c5) at (11, 1.5) {$a_5$};
\node[boxBasis] (c6) at (12, 1.5) {$a_6$};
\node[boxBasis] (c7) at (13, 1.5) {$a_7$};  
\node[boxBasis] (c8) at (14, 1.5) {$a_8$};
\node[fill=red!10] () at (14.9, 1.5) {$\cdots$};
\node[fill=red!10] () at (15.4, 1.5) {$\cdots$};
\node[fill=red!10] () at (15.9, 1.5) {$\cdots$};

\draw[->] (b3) to (c3) node[ ] {};
\draw[->] (b4) to (c4) node[ ] {};
\draw[->] (b5) to (c5) node[ ] {};
\draw[->] (b6) to (c6) node[ ] {};
\draw[->] (b7) to (c7) node[ ] {};
\draw[->] (b8) to (c8) node[ ] {};

\node[fill=gray!10] () at (11, 0.8) {Multiplication by~2 means that iterated applications of~$\mathcal{D}$ progressively shift the fractional part};
\node[fill=gray!10] () at (11, 0.2) {of the binary representation of~$\alpha_k$; the integral part is always~0 because of the mod operation.};

\node[myboxA] (ex0) at (4.5, -0.6) {$\alpha_{k+6} = \mathcal{D}^6(\alpha_k)$};
\node[myboxLeading] (ex0) at (7.15, -0.6) {$a_6\, \text{mod}\, 1 \equiv 0.$};
\node[boxBasis] (d7) at (9, -0.6) {$a_7$};  
\node[boxBasis] (d8) at (10, -0.6) {$a_8$};
\node[fill=red!10] () at (10.9, -0.6) {$\cdots$};
\node[fill=red!10] () at (11.4, -0.6) {$\cdots$};
\node[fill=red!10] () at (11.9, -0.6) {$\cdots$};

\node[myboxA] (ex0) at (4.5, -1.5) {$\alpha_{k+7} = \mathcal{D}^7(\alpha_k)$};
\node[myboxLeading] (ex0) at (7.15, -1.5) {$a_7\, \text{mod}\, 1 \equiv 0.$};
\node[boxBasis] (d7) at (9, -1.5) {$a_8$};  
\node[fill=red!10] () at (9.9, -1.5) {$\cdots$};
\node[fill=red!10] () at (10.4, -1.5) {$\cdots$};
\node[fill=red!10] () at (10.9, -1.5) {$\cdots$};

\node[myboxA] (ex0) at (4.5, -2.4) {$\alpha_{k+8} = \mathcal{D}^8(\alpha_k)$};
\node[myboxLeading] (ex0) at (7.15, -2.4) {$a_8\, \text{mod}\, 1 \equiv 0.$};

\draw[->] (d8) to (d7) node[ ] {};

\node[fill=red!10] () at (8.85, -2.4) {$\cdots$};
\node[fill=red!10] () at (9.35, -2.4) {$\cdots$};
\node[fill=red!10] () at (9.85, -2.4) {$\cdots$};

\end{tikzpicture} \end{center} 

\noindent In addition to clarifying the bit-shift property of the dyadic transformation, this visualization nicely brings to light the fundamental mechanism responsible for the onset of chaotic dynamics: each iteration of~$\mathcal{D}$ leads to the loss of~1~bit of information.  After~$\tau$ iterations all the significant bits of~$\alpha_k$ have been lost and we are left with an infinite sequence of random bits.  In other words, the time evolution of~$\alpha_k$ depends very strongly on its least significant bits: a hallmark of chaos known as sensitivity to initial conditions.

\newpage

\paragraph{Encoder/decoder strategies.}  Let us consider a dataset~$\mathcal{X} = [ x_0, \cdots , x_n ]$ composed of a sequence of samples whose values have been normalized to lie in the unit interval. Any data source, regardless of the underlying dimensionality or modality, can be collapsed into such a list of numerical values and min-max normalized so that whatever pre-processing procedure one may need to apply to the raw data does not impose any restrictions on our approach (as demonstrated before by the illustrative examples for scatter plots, sound, images...). \\

\noindent Already, combining the fixed-point binary representation of real numbers along with the dyadic transformation allows us to construct a straightforward encoding/decoding strategy that is able to reproduce all the samples of~$\mathcal{X}$ up to an arbitrary degree of precision using a single parameter in~$\mathbb{R}$ (section~\ref{lab:naive}).  Because of its lack of sophistication, this strategy won't be able to explain the elegant expression~$f_\alpha$ displayed at the very beginning of the paper.  Nevertheless, it will serve as a conceptual scaffolding upon which our second strategy will be heavily based.  Uncovering the origin of~$f_\alpha$ will require the introduction a few more theoretical tools from chaos theory: logistic maps and the concept of topological conjugacy.  Blending these new ingredients on top of the blueprint laid out in the first construction will finally reveal how the differentiable model~$f_\alpha$ comes about (section~\ref{secondStrategy}).

\section{Building the scaffolding}
\label{lab:naive}

The idea is almost embarrassingly simple...  Start by encoding all the values of~$\mathcal{X}$ into a long binary string and convert it to its decimal representation.  Use this number~$\alpha_0 \in \mathbb{R}$ as the initial condition for the dyadic transformation.  Because of its bit-shift property, iterated applications of~$\mathcal{D}$ can be used to decode one-by-one all the components of~$\mathcal{X}$.

\subsection{Construction of the initial condition}
\label{naive::init}

We start by converting all the values~$[x_0, x_1, x_2, \cdots , x_n]$ of~$\mathcal{X}$ into their binary representation truncated to the first~$\tau$ significant bits. \\
\begin{center} \begin{tikzpicture}

\node[] (y0) at (4, 0) {$x_0 \approx x^0_\text{bin} = $ };
\node[] (y1) at (4, -1) {$x_1 \approx x^1_\text{bin} = $ };
\node[] (y2) at (4, -2) {$x_2 \approx x^2_\text{bin} = $ };
\node[] (yn) at (4, -4) {$x_n \approx x^n_\text{bin} = $ };

\node[myboxFst] (a1) at (5.3, 0) { \phantom{0} };
\node[myboxFst] (a2) at (5.8, 0) {\phantom{0} };
\node[myboxFst] (a3) at (6.3, 0) {\phantom{0}  };
\node[myboxFst] (a4) at (6.8, 0) {\phantom{0} };
\node[myboxFst] (a5) at (7.3, 0) {\phantom{0}  };
\node[myboxFst] (a6) at (7.8, 0) {\phantom{0} };
\node[myboxFst] (a7) at (8.3, 0) {\phantom{0} };
\node[myboxFst] (a8) at (8.8, 0) { \phantom{0}};

\node[myboxSec] (b1) at (5.3, -1) { \phantom{0} };
\node[myboxSec] (b2) at (5.8, -1) { \phantom{0}};
\node[myboxSec] (b3) at (6.3, -1) { \phantom{0} };
\node[myboxSec] (b4) at (6.8, -1) {\phantom{0} };
\node[myboxSec] (b5) at (7.3, -1) {\phantom{0}  };
\node[myboxSec] (b6) at (7.8, -1) { \phantom{0}};
\node[myboxSec] (b7) at (8.3, -1) { \phantom{0}};
\node[myboxSec] (b8) at (8.8, -1) {\phantom{0} };

\node[myboxLst] (c1) at (5.3, -2) { \phantom{0} };
\node[myboxLst] (c2) at (5.8, -2) {\phantom{0} };
\node[myboxLst] (c3) at (6.3, -2) {\phantom{0}  };
\node[myboxLst] (c4) at (6.8, -2) { \phantom{0}};
\node[myboxLst] (c5) at (7.3, -2) {\phantom{0}  };
\node[myboxLst] (c6) at (7.8, -2) { \phantom{0}};
\node[myboxLst] (c7) at (8.3, -2) { \phantom{0}};
\node[myboxLst] (c8) at (8.8, -2) {\phantom{0} };

\node[] () at (3.8, -2.8) {$\vdots$ };
\node[] () at (3.8, -3.2) {$\vdots$ };

\node[myboxLstFinal] (d1) at (5.3, -4) { \phantom{0} };
\node[myboxLstFinal] (d2) at (5.8, -4) {\phantom{0} };
\node[myboxLstFinal] (d3) at (6.3, -4) {\phantom{0}  };
\node[myboxLstFinal] (d4) at (6.8, -4) { \phantom{0}};
\node[myboxLstFinal] (d5) at (7.3, -4) {\phantom{0}  };
\node[myboxLstFinal] (d6) at (7.8, -4) { \phantom{0}};
\node[myboxLstFinal] (d7) at (8.3, -4) { \phantom{0}};
\node[myboxLstFinal] (d8) at (8.8, -4) {\phantom{0} };

\draw [decorate,decoration={brace,amplitude=10pt}]
(a8.north east) -- (d8.south east) node [align=left, black,midway,xshift=4cm] 
{Convert all samples to fixed-point binary \\ representation with~$\tau$ bits of precision};

\end{tikzpicture} \end{center} 

\noindent Next, we concatenate all the samples together so that the entire dataset is ``flattened'' into a long binary string composed of~$\sim n \tau$ bits.  Notice how we associate the binary coefficients with weights of decreasing significance according to their order of appearance in~$\mathcal{X}$. \\ \\
\noindent \makebox[\textwidth][c]{\begin{tikzpicture}   

\node[] (alpha) at (0.2, 0) {$\alpha_{0} = $ };

\node[myboxFst] (a1) at (1, 0) { \phantom{0} };
\node[myboxFst] (a2) at (1.5, 0) { \phantom{0}  };
\node[myboxFst] (a3) at (2, 0) {\phantom{0} };
\node[myboxFst] (a4) at (2.5, 0) {\phantom{0} };
\node[myboxFst] (a5) at (3, 0) { \phantom{0} };
\node[myboxFst] (a6) at (3.5, 0) { \phantom{0} };
\node[myboxFst] (a7) at (4, 0) {\phantom{0} };
\node[myboxFst] (a8) at (4.5, 0) {\phantom{0} };
\node[myboxSec] (b1) at (5, 0) { \phantom{0} };
\node[myboxSec] (b2) at (5.5, 0) { \phantom{0}};
\node[myboxSec] (b3) at (6, 0) {\phantom{0}  };
\node[myboxSec] (b4) at (6.5, 0) {\phantom{0} };
\node[myboxSec] (b5) at (7, 0) {\phantom{0}  };
\node[myboxSec] (b6) at (7.5, 0) {\phantom{0} };
\node[myboxSec] (b7) at (8, 0) {\phantom{0} };
\node[myboxSec] (b8) at (8.5, 0) {\phantom{0} };

\node[myboxLst] (l1) at (9, 0) {\phantom{0} };
\node[myboxLst] (l2) at (9.5, 0) {\phantom{0} };
\node[myboxLst] (l3) at (10, 0) {\phantom{0} };
\node[myboxLst] (l4) at (10.5, 0) {\phantom{0} };
\node[myboxLst] (l5) at (11, 0) {\phantom{0} };
\node[myboxLst] (l6) at (11.5, 0) {\phantom{0} };
\node[myboxLst] (l7) at (12, 0) {\phantom{0} };
\node[myboxLst] (l8) at (12.5, 0) {\phantom{0} };

\node[minimum width=0.5cm] (bDots) at (13.15, 0) {  $\cdots$ };
\node[minimum width=0.5cm] (bDots2) at (13.65, 0) {  $\cdots$ };
\node[minimum width=0.5cm] (bDots3) at (14.15, 0) {  $\cdots$ };

\node[myboxLstFinal] (y1) at (14.8, 0) {\phantom{0} };
\node[myboxLstFinal] (y2) at (15.3, 0) {\phantom{0} };
\node[myboxLstFinal] (y3) at (15.8, 0) {\phantom{0} };
\node[myboxLstFinal] (y4) at (16.3, 0) {\phantom{0} };
\node[myboxLstFinal] (y5) at (16.8, 0) {\phantom{0} };
\node[myboxLstFinal] (y6) at (17.3, 0) {\phantom{0} };
\node[myboxLstFinal] (y7) at (17.8, 0) {\phantom{0} };
\node[myboxLstFinal] (y8) at (18.3, 0) {\phantom{0} };

\node[myboxFst, fill=gray!10] (basis1) at (1, -1) {$1/2$};
\node[myboxFst, fill=gray!10] (basis2) at (4.5, -1) {$1/2^\tau$};
\node[myboxSec, fill=gray!10] (basis3) at (8.5, -1) {$1/2^{2\tau}$ };
\node[myboxSec, fill=gray!10] (basis4) at (12.5, -1) {$1/2^{3\tau}$ };
\node[myboxSec, fill=gray!10] (basis5) at (14.8, -1) {$1/2^{n\tau}$ };
\node[myboxSec, fill=gray!10] (basis6) at (18.3, -1) {$1/2^{(n+1)\tau}$ };

\draw [decorate,decoration={brace,amplitude=10pt},xshift=60pt]
(y1.north west) -- (y8.north east) node [black,midway,yshift=0.3cm,above] 
{$x^n_\text{bin}$};

\draw[->, dashed] (a1.south) to (basis1.north) node[ ] { };
\draw[->, dashed] (a8.south) to (basis2.north) node[ ] { };
\draw[->, dashed] (b8.south) to (basis3.north) node[ ] { };
\draw[->, dashed] (l8.south) to (basis4.north) node[ ] { };
\draw[->, dashed] (y1.south) to (basis5.north) node[ ] { };
\draw[->, dashed] (y8.south) to (basis6.north) node[ ] { };

\draw [decorate,decoration={brace,amplitude=10pt},xshift=60pt]
(a1.north west) -- (a8.north east) node [black,midway,yshift=0.3cm,above] 
{$x^0_\text{bin}$};

\draw [decorate,decoration={brace,amplitude=10pt},xshift=60pt]
(b1.north west) -- (b8.north east) node [black,midway,yshift=0.3cm,above] 
{$x^1_\text{bin}$};

\draw [decorate,decoration={brace,amplitude=10pt},xshift=60pt]
(l1.north west) -- (l8.north east) node [black,midway,yshift=0.3cm,above] 
{$x^2_\text{bin}$};

\draw [decorate,decoration={mirror,brace,amplitude=10pt},xshift=60pt] (basis1.south) -- (basis2.south) node [black,midway,yshift=-0.3cm,below] {$\alpha_0 \approx x^0_\text{bin}$};

\end{tikzpicture}} \\

\noindent Thanks to this encoding scheme, the bits corresponding to~$x^0_\text{bin}$ occupy the first~$\tau$ dominant weights ensuring that~$\alpha_0$ is already a good approximation to the value of the first sample:
\begin{equation*}
\highlight[black!10]{\alpha_0 \approx x_0 \quad \text{with error bound} \quad \left| \alpha_0 - x_0  \right| < 1/2^\tau}
\end{equation*}
The error bound is explained in the introductory paragraph on fixed-point binary representations.

\subsection{Decimal precision requirements}
\label{sec::decimalPrecision}

Converting~$\alpha_0 \in [0,1]$ to its equivalent decimal representation is achieved by evaluating the expression:
\begin{equation*}
\highlight[black!10]{\alpha_0 = \sum_{i=1}^{(n+1)\tau} \dfrac{\alpha_i}{2^i} } = \left(  \highlight[yellow!10]{\sum_{i=1}^\tau}  + \highlight[green!10]{\sum_{i=\tau+1}^{2 \tau}}  + \highlight[pink!10]{\sum_{n=2\tau+1}^{3 \tau}}  + \cdots + \highlight[blue!10]{\sum_{n=n\tau + 1}^{(n+1)\tau}} \right) \dfrac{\alpha_i}{2^i} 
\end{equation*}
Conceptually, this summation can be partitioned into blocks which are color-coded according to the level of significance they represent in the original binary representation of~$\alpha_0$. Each block encodes the value~$x_j$ of a particular sample corrected by a multiplicative factor determined by the order in which sample~$j$ appears in the dataset~$\mathcal{X} = [x_0, x_1, x_2, \cdots , x_n]$. \\

\noindent Since all further arithmetic operations will exclusively be performed in decimal representation, it is important to answer the following question: How many digits of decimal precision does one need to keep in order to have the capacity to represent all the information contained in the original construction of~$\alpha_0$ as a long binary string?  Let us denote by~$p_\text{bin}$ the level of binary precision (number of bits) targeted.  This means that we aim to describe numbers with a numerical precision of at least~$1 / 2^{\,p_\text{bin}}$.  Therefore the required level of decimal precision~$p_\text{dec}$ is determined by:
\begin{equation*}
\dfrac{1}{{10}^{\,p_\text{dec}}}  = \dfrac{1}{2^{\,p_\text{bin}}} \quad 
\Longrightarrow \quad p_\text{dec} = \left( \dfrac{\log 2}{\log 10} \approx 0.3 \right) p_\text{bin}
\end{equation*}

\noindent Unsurprisingly, decimal representation requires less digits than would otherwise be necessary in binary to achieve the same level of numerical precision.  Intuitively, this can be attributed to the fact the decimal representation of a real number can be expanded over a larger set of base digits~$\{ 0, \cdots , 9 \}$ instead of just~$\{ 0,1 \}$ in binary form. \\

\noindent With typical numbers such as~$n=1000$ data points and an accuracy of~$\tau=8$ bits of precision for each sample, we need to faithfully encode~$p_\text{bin} \approx 8000$ bits of information which corresponds to~$p_\text{dec} \approx 2400$ significant decimal digits.  Obviously, this level of precision goes way beyond the traditional IEEE floating-point standard and other common data types offered by host programming languages.  Therefore, actual implementation of the algorithm requires the use of external libraries to perform arbitrary-precision arithmetic operations. 

\subsection{Decoding using the dyadic transformation} 

\noindent Going back to the definition of~$\mathcal{D}$ in~eq.(\ref{eq:dyadic}), it is easy to see that multiple iterated applications of the dyadic transformation~$k\tau$ times on the initial condition~$\alpha_0$ leads to a new value~$\alpha_k$ which is given by:
\begin{equation}
\alpha_k \equiv \underbrace{\left( \mathcal{D} \circ \cdots \circ \mathcal{D} \right)}_{k\tau \text{ times}} \, (\alpha_0) = \mathcal{D}^{k\tau} (\alpha_0) \quad \Longrightarrow \quad \highlight[black!10]{ \alpha_k = 2^{k\tau} \, \alpha_0 \, \text{mod} \,\, 1 \quad ; \quad k \in \{ 0, \cdots , n \} }
\label{eq:exactSol}
\end{equation}
As will be demonstrated below, this expression for~$\alpha_k$ is indeed able to reproduce all the datapoints of~$\mathcal{X}$.  \\

\noindent With~$k=0$, we have seen that~$\alpha_0$ is already a good approximation to the first sample of our dataset.  In order to see how the remaining datapoints can also be extracted using~$\mathcal{D}$, it is helpful to mentally go back to the binary representation in which the dyadic transformation can be seen as a simple leftwise bit-shift.  Note that this logic is only presented {\bf as a visual guide} and that~$\alpha$'s are calculated directly in decimal basis.  Accordingly, let us start by looking at~$k=1$ so that the first iterate~$\alpha_1$ corresponds to: \\ \\
\noindent \makebox[\textwidth][c]{\begin{tikzpicture}   

\node[] (alpha) at (-0.6, 0) {$\alpha_1 = \mathcal{D}^\tau \left( \alpha_0 \right) = $ };

\node[myboxSec] (a1) at (1, 0) { \phantom{0} };
\node[myboxSec] (a2) at (1.5, 0) { \phantom{0}  };
\node[myboxSec] (a3) at (2, 0) {\phantom{0} };
\node[myboxSec] (a4) at (2.5, 0) {\phantom{0} };
\node[myboxSec] (a5) at (3, 0) { \phantom{0} };
\node[myboxSec] (a6) at (3.5, 0) { \phantom{0} };
\node[myboxSec] (a7) at (4, 0) {\phantom{0} };
\node[myboxSec] (a8) at (4.5, 0) {\phantom{0} };

\node[myboxLst] (b1) at (5, 0) { \phantom{0} };
\node[myboxLst] (b2) at (5.5, 0) { \phantom{0}};
\node[myboxLst] (b3) at (6, 0) {\phantom{0}  };
\node[myboxLst] (b4) at (6.5, 0) {\phantom{0} };
\node[myboxLst] (b5) at (7, 0) {\phantom{0}  };
\node[myboxLst] (b6) at (7.5, 0) {\phantom{0} };
\node[myboxLst] (b7) at (8, 0) {\phantom{0} };
\node[myboxLst] (b8) at (8.5, 0) {\phantom{0} };

\node[minimum width=0.5cm] () at (9.15, 0) {  $\cdots$ };
\node[minimum width=0.5cm] () at (9.65, 0) {  $\cdots$ };
\node[minimum width=0.5cm] () at (10.15, 0) {  $\cdots$ };

\node[myboxLstFinal] (y1) at (10.8, 0) {\phantom{0} };
\node[myboxLstFinal] (y2) at (11.3, 0) {\phantom{0} };
\node[myboxLstFinal] (y3) at (11.8, 0) {\phantom{0} };
\node[myboxLstFinal] (y4) at (12.3, 0) {\phantom{0} };
\node[myboxLstFinal] (y5) at (12.8, 0) {\phantom{0} };
\node[myboxLstFinal] (y6) at (13.3, 0) {\phantom{0} };
\node[myboxLstFinal] (y7) at (13.8, 0) {\phantom{0} };
\node[myboxLstFinal] (y8) at (14.3, 0) {\phantom{0} };

\node[myboxFst, fill=gray!10] (basis1) at (1, -1) {$1/2$};
\node[myboxFst, fill=gray!10] (basis2) at (4.5, -1) {$1/2^\tau$};
\node[myboxSec, fill=gray!10] (basis3) at (8.5, -1) {$1/2^{2\tau}$ };
\node[myboxSec, fill=gray!10] (basis5) at (10.8, -1) {$1/2^{(n-1)\tau}$ };
\node[myboxSec, fill=gray!10] (basis6) at (14.3, -1) {$1/2^{n\tau}$ };

\draw[->, dashed] (a1.south) to (basis1.north) node[ ] { };
\draw[->, dashed] (a8.south) to (basis2.north) node[ ] { };
\draw[->, dashed] (b8.south) to (basis3.north) node[ ] { };
\draw[->, dashed] (y1.south) to (basis5.north) node[ ] { };
\draw[->, dashed] (y8.south) to (basis6.north) node[ ] { };

\draw [decorate,decoration={brace,amplitude=10pt},xshift=60pt]
(a1.north west) -- (a8.north east) node [black,midway,yshift=0.3cm,above] 
{$x^1_\text{bin}$};

\draw [decorate,decoration={brace,amplitude=10pt},xshift=60pt]
(b1.north west) -- (b8.north east) node [black,midway,yshift=0.3cm,above] 
{$x^2_\text{bin}$};

\draw [decorate,decoration={brace,amplitude=10pt},xshift=60pt]
(y1.north west) -- (y8.north east) node [black,midway,yshift=0.3cm,above] 
{$x^n_\text{bin}$};

\draw [decorate,decoration={mirror,brace,amplitude=10pt},xshift=60pt] (basis1.south) -- (basis2.south) node [black,midway,yshift=-0.3cm,below] {$\alpha_1 \approx x^1_\text{bin}$};

\end{tikzpicture}}

\noindent showing that the second sample is recovered by~$\tau$ successive applications of~$\mathcal{D}$ on our carefully crafted initial condition:
\begin{equation*}
\highlight[black!10]{\alpha_1 = 2^\tau \alpha_0 \, \text{mod} \,\, 1 \,\, \Longrightarrow \,\, \alpha_1 \approx x_1 \,\,\, \text{with error bound} \,\,\, \left| \alpha_1 - x_1  \right| < 1/2^\tau}
\end{equation*} 

\newpage

\noindent Moving on to~$k=2$, we have: \\ \\
\noindent \begin{tikzpicture}   

\node[] (alpha) at (-0.7, 0) {$\alpha_{2} = \mathcal{D}^{2\tau} \left( \alpha_0 \right) = $ };

\node[myboxLst] (a1) at (1, 0) { \phantom{0} };
\node[myboxLst] (a2) at (1.5, 0) { \phantom{0}  };
\node[myboxLst] (a3) at (2, 0) {\phantom{0} };
\node[myboxLst] (a4) at (2.5, 0) {\phantom{0} };
\node[myboxLst] (a5) at (3, 0) { \phantom{0} };
\node[myboxLst] (a6) at (3.5, 0) { \phantom{0} };
\node[myboxLst] (a7) at (4, 0) {\phantom{0} };
\node[myboxLst] (a8) at (4.5, 0) {\phantom{0} };

\node[minimum width=0.5cm] () at (5.15, 0) {  $\cdots$ };
\node[minimum width=0.5cm] () at (5.65, 0) {  $\cdots$ };
\node[minimum width=0.5cm] () at (6.15, 0) {  $\cdots$ };

\node[myboxLstFinal] (y1) at (6.8, 0) {\phantom{0} };
\node[myboxLstFinal] (y2) at (7.3, 0) {\phantom{0} };
\node[myboxLstFinal] (y3) at (7.8, 0) {\phantom{0} };
\node[myboxLstFinal] (y4) at (8.3, 0) {\phantom{0} };
\node[myboxLstFinal] (y5) at (8.8, 0) {\phantom{0} };
\node[myboxLstFinal] (y6) at (9.3, 0) {\phantom{0} };
\node[myboxLstFinal] (y7) at (9.8, 0) {\phantom{0} };
\node[myboxLstFinal] (y8) at (10.3, 0) {\phantom{0} };

\node[myboxFst, fill=gray!10] (basis1) at (1, -1) {$1/2$};
\node[myboxFst, fill=gray!10] (basis2) at (4.5, -1) {$1/2^\tau$};
\node[myboxSec, fill=gray!10] (basis5) at (6.8, -1) {$1/2^{(n-2)\tau}$ };
\node[myboxSec, fill=gray!10] (basis6) at (10.3, -1) {$1/2^{(n-1)\tau}$ };

\draw [decorate,decoration={brace,amplitude=10pt},xshift=60pt] (y1.north west) -- (y8.north east) node [black,midway,yshift=0.3cm,above] {$x^n_\text{bin}$};

\draw [decorate,decoration={brace,amplitude=10pt},xshift=60pt] (a1.north west) -- (a8.north east) node [black,midway,yshift=0.3cm,above] {$x^2_\text{bin}$};

\draw [decorate,decoration={mirror,brace,amplitude=10pt},xshift=60pt] (basis1.south) -- (basis2.south) node [black,midway,yshift=-0.3cm,below] {$\alpha_{2} \approx x^2_\text{bin}$};

\draw[->, dashed] (a1.south) to (basis1.north) node[ ] { };
\draw[->, dashed] (a8.south) to (basis2.north) node[ ] { };
\draw[->, dashed] (y1.south) to (basis5.north) node[ ] { };
\draw[->, dashed] (y8.south) to (basis6.north) node[ ] { };

\end{tikzpicture} \\ 

\noindent confirming that the third sample of~$\mathcal{X}$ is indeed reproduced:
\begin{equation*}
\highlight[black!10]{\alpha_2 = 2^{2\tau} \alpha_0 \, \text{mod} \,\, 1 \,\, \Longrightarrow \,\, \alpha_2 \approx x_2 \,\,\, \text{with error bound} \,\,\, \left| \alpha_2 - x_2  \right| < 1/2^\tau}
\end{equation*}
Generalizing, it is clear that iterating the dyadic transformation~$n\tau$ times removes away all the bits of~$\alpha_0$ leaving only its (initally) most subdominant contribution (now dominant) which was constructed as the sequence of~$\tau$ bits encoding the last sample: \\

\noindent \begin{tikzpicture}   

\node[] (alpha) at (-0.8, 0) {$\alpha_n = \mathcal{D}^{n\tau} \left( \alpha_\text{bin} \right) = $ };

\node[myboxLstFinal] (a1) at (1, 0) { \phantom{0} };
\node[myboxLstFinal] (a2) at (1.5, 0) { \phantom{0}  };
\node[myboxLstFinal] (a3) at (2, 0) {\phantom{0} };
\node[myboxLstFinal] (a4) at (2.5, 0) {\phantom{0} };
\node[myboxLstFinal] (a5) at (3, 0) { \phantom{0} };
\node[myboxLstFinal] (a6) at (3.5, 0) { \phantom{0} };
\node[myboxLstFinal] (a7) at (4, 0) {\phantom{0} };
\node[myboxLstFinal] (a8) at (4.5, 0) {\phantom{0} };

\node[myboxFst, fill=gray!10] (basis1) at (1, -1) {$1/2$};
\node[myboxFst, fill=gray!10] (basis2) at (4.5, -1) {$1/2^\tau$};

\draw [decorate,decoration={brace,amplitude=10pt},xshift=60pt] (a1.north west) -- (a8.north east) node [black,midway,yshift=0.3cm,above] {$x^n_\text{bin}$};

\draw[->, dashed] (a1.south) to (basis1.north) node[ ] { };
\draw[->, dashed] (a8.south) to (basis2.north) node[ ] { };

\end{tikzpicture} \\

\noindent thereby completing our demonstration that the entire dataset~$\mathcal{X}$ can be decoded by~eq.(\ref{eq:exactSol}):
\begin{equation*}
\highlight[black!10]{\alpha_n = 2^{n\tau} \alpha_0 \, \text{mod} \,\, 1 \,\, \Longrightarrow \,\, \alpha_n \approx x_n \,\,\, \text{with error bound} \,\,\, \left| \alpha_n - x_n  \right| < 1/2^\tau}    
\end{equation*}

\subsection{Summary in code}
\label{codeSummary::1}

\noindent Let us finish this section by showing how the algorithm described above can be implemented very easily in just a few lines of code. {\it (Helper functions to convert between binary and decimal representations, using a default value of~$\tau=8$, as well as the construction of a dataset~$\mathcal{X}$ composed of~50 random numbers in the unit interval are provided in appendix~\ref{code::appendix}.)} \\

\noindent The first step consists in looping over the dataset to approximate the decimal values of all of its samples with their~$\tau$-bit binary representations.  This transformation of~$\mathcal{X} = [x_0, x_1, x_2, \cdots, x_n] \approx [x_\text{bin}^0, x_\text{bin}^1,x_\text{bin}^2,\cdots,x_\text{bin}^n ]$ is achieved by using the helper function~\texttt{decimalToBinary} (which internally invokes the dyadic transformation implemented as~\texttt{dyadicMap} below). Appending these~$\tau$-long binary strings all together by order of appearance in~$\mathcal{X}$ (denoted in code as~\texttt{xs}) leads to the initial condition~$\alpha_0$ (denoted in code as~\texttt{binaryInitial}). \\
\begin{lstlisting}[language=Python, caption=]
# xs: List[float] is the dataset instantiated here as a list of random values in [0, 1] as defined the Appendix.

dyadicMap = lambda x : (2 * x) % 1

binaryInitial = ''.join(map(decimalToBinary, xs))

print('binaryInitial = %s\n' % binaryInitial)

binaryInitial = 1000110010110111100110101000101101101100101001010111000011100100111101100110
                0010110010101000011110010001111011000001001000010110000001011101010111000111
                1101111011111010110011000111011011000111000111101010001100100100111100011000
                0101011010100100001111000110011101001001000100000100100111101001110010011101
                1111000110101110010111000110111110110010000011111010101010101011001101010010
                00010101000001011101

\end{lstlisting}

\vspace{0.3cm}

\noindent Since~$\mathcal{X}\equiv$~\texttt{xs} is composed of~50 samples represented in binary with~$\tau=8$ bits of precision, we can verify that the initial condition~\texttt{binaryInitial} requires~$50 \times 8 = 400$ bits of precision. 

\newpage

\noindent Converting this initial condition from its binary form into its corresponding decimal representation is achieved by invoking the~\texttt{binaryToDecimal} helper function. \\
\begin{lstlisting}[language=Python, caption=]
decimalInitial = binaryToDecimal(binaryInitial)

print('decimalInitial = %s' % decimalInitial)

decimalInitial = 0.5496765699760055703169202362353308700123406974106022231112441515212066847
                  3990606828049163011967627260871299016527460727972        

\end{lstlisting}

\vspace{0.3cm}

\noindent As can be verified from the complete value of~$\alpha_0 \equiv$~\texttt{decimalInitial}, one needs to keep about~$\approx 120$ decimal digits of precision.  Next, we implement the decoding function~$\alpha_k \equiv$~\texttt{dyadicDecoder}. \\
\begin{lstlisting}[language=Python, caption=]
dyadicDecoder = lambda k: (2 ** (k * tau) * decimalInitial) % 1
\end{lstlisting}

\vspace{0.3cm}

\noindent It is important to recognize that the multiplication between~$\alpha_0$ and the~$k$-dependent exponential pre-factor should be understood in the context of ``overloading''.  In particular, this means that the implementation is dispatched to an external library that handles arithmetic operations on arbitrary-precision objects such as~\texttt{decimalInitial} (instead of being handled by the native implementations on common data types offered by the host programming language). \\

\noindent Now that we have constructed the initial condition and the decoding function, all the values of the original dataset~$\mathcal{X}$ can be generated (within a level of precision determined by~$\tau$) one-by-one by applying~\texttt{dyadicDecoder} on a range of integer-valued arguments~$k \in [0, \cdots , n]$. Even though arbitrary-precision numbers were necessary to carry out accurate arithmetic operations on~\texttt{decimalInitial}, the components of the resulting list denoted as~\texttt{decodedValues} can be downcast to usual floating-point numbers using the built-in~\texttt{float} method. \\
\begin{lstlisting}[language=Python, caption=]
n = len(xs)
decodedValues = [float(dyadicDecoder(_)) for _ in range(n)]
\end{lstlisting}

\vspace{0.2cm}

\noindent As a sanity check, it is interesting to check that the difference between the values~$[\alpha_0, \alpha_1, \alpha_2, \cdots, \alpha_n]$ produced by our encoding/decoding strategy and the true values from~$\mathcal{X}=[x_0, x_1, x_2, \cdots , x_n]$ satisfy the theoretical error bound~$\left| \alpha_j - x_j  \right| < 1/2^\tau$ captured by~\texttt{normalizedErrors} in the code fragment below.  \\
\begin{lstlisting}[language=Python, caption=]
maxError = 1 / 2 ** tau
normalizedErrors = [abs(decodedValue - dataPoint) / maxError 
                    for decodedValue, dataPoint in zip(decodedValues, xs)]
\end{lstlisting}

\begin{figure}[hb!]
\centering
\begin{subfigure}{.5\textwidth}
\centering
\includegraphics[width=\linewidth]{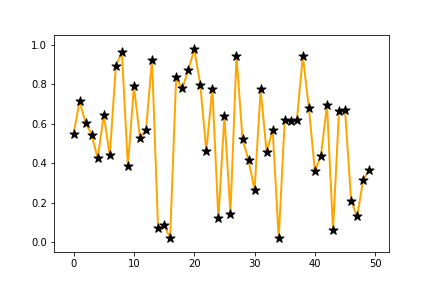}
\end{subfigure}%
\begin{subfigure}{.5\textwidth}
\centering
\includegraphics[width=\linewidth]{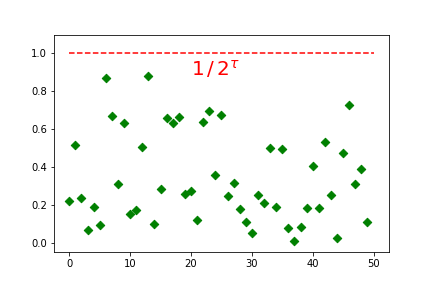}
\end{subfigure}
\caption{{\bf Left)} Comparison between the decoded values (black star markers) against the original dataset~$\mathcal{X}$ (thick orange line). {\bf Right)} Normalized error showing that the gap between decoded values and the original values never strays more than~$1/2^\tau$ away (represented as the red horizontal dashed line) confirming the error bound derived in the main text.}
\label{fig:naivePics}
\end{figure}

\newpage

\section{Applying makeup to the scaffolding}
\label{secondStrategy}

We refer to the strategy described in the previous section as {\it ``building the scaffolding''}.  The reason for this choice of terminology is that it feels like the decoding function leaves too much of the bare mechanics exposed.  Most objectionable, it relies in plain view on the dyadic transformation and by implication on its bit-shift property.  Moreover, the somewhat inelegant modulo operation with its lack of continuity may stand out as a ``hair in the soup'' in an age where \href{https://en.wikipedia.org/wiki/Differentiable\_programming}{\bf differentiable programming} is all the rage. \\

\noindent The purpose of this section is to justify the more aesthetically pleasing formulation~$f_\alpha(x) = \sin^2 (2^{x\tau} \arcsin{\sqrt{\alpha}})$ whose accomplishments have already been documented in the introduction as a replacement to the crude decoding function,~eq.(\ref{eq:exactSol}), that we currently have.  To do this, we need to introduce two well-known concepts in the context of dynamical systems: the logistic map and the notion of topological conjugacy.  It is worth emphasizing that these ingredients will be used simply as a kind of ``makeup'' on top of an otherwise mostly identical logic to the one presented in the previous section.

\subsection{Logistic map}

Originally introduced as a discrete-time model for population demographics, the logistic map~$\mathcal{L}$ is a polynomial mapping between a real-valued~$z_k \in [0,1]$ and its next iterate~$z_{k+1}$ defined by:
\begin{equation}
z_{k+1} \equiv \mathcal{L} (z_k) = r \, z_k (1 - z_k)
\label{eq:logisiticMap}
\end{equation}
Note that we will restrict ourselves to the special case where~$r=4$.  Even if not particularly useful in biology, it turns out that the logistic map exhibits an unexpected degree of complexity~\cite{robertMay}.  In fact, it is often used nowadays as the \href{https://en.wikipedia.org/wiki/Logistic\_map}{archetypal example} of how chaotic behavior may arise from simple dynamical equations. \\

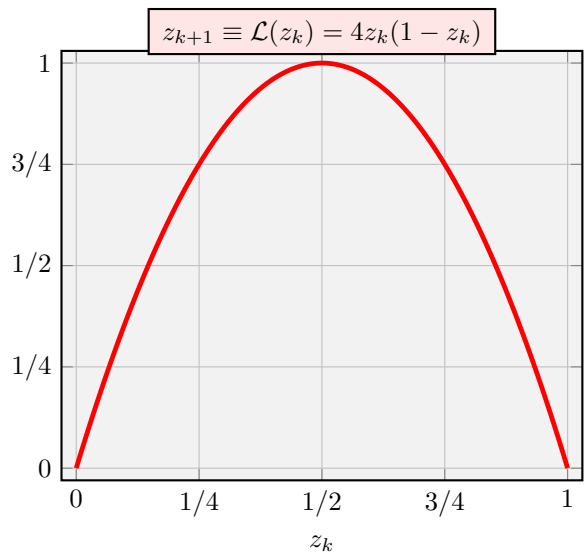
\begin{figure}[hb!]
\begin{center}
\begin{tikzpicture}   
\begin{axis}[xmin=-0.03, xmax=1.03, ymin=-0.03, ymax=1.03, legend pos=north west, legend style={empty legend, fill=red!10, at={(0.5,1.1)},anchor=north}, axis background/.style={fill=gray!10}, xtick={0, 0.25, 0.5, 0.75, 1}, xticklabels={0, 1/4, 1/2, 3/4, 1},
ytick={0, 0.25, 0.5, 0.75, 1}, yticklabels={0, 1/4, 1/2, 3/4, 1}, thick, minor x tick num=0, minor y tick num=0, grid=both, xlabel=$z_k$]
\addplot[samples=1000,domain=0:1, color=red, line width=1.8pt]{4 * x * (1-x)};
\addlegendentry{$z_{k+1} \equiv \mathcal{L} (z_k) = 4 z_k (1-z_k)$};
\end{axis}
\end{tikzpicture}
\end{center} 
\caption{Graphical illustration of the logistic map~$\mathcal{L}$ as defined in~eq.(\ref{eq:logisiticMap}) in the special case where~$r=4$. To be compared with the non-differentiable dyadic transformation~$\mathcal{D}$ shown in~Figure~\ref{fig:dyadic}.}
\label{fig:logistic}
\end{figure}

\noindent A quick visual comparison of~$\mathcal{L}$ (as depicted in~Figure~\ref{fig:logistic}) with the dyadic transformation~$\mathcal{D}$ (illustrated previously in~Figure~\ref{fig:dyadic}) suggests that both maps do not have much in common with each other...  For example, it is clear that~$\mathcal{L}$ is continuous and differentiable over the unit interval; smoothness guarantees that stand in stark contrast to the sawtooth shape associated with~$\mathcal{D}$. \\

\noindent At this point you may ask: well, what does this logistic map business have to do with anything we discussed so far? \\

\subsection{Toplogical conjugacy}
\label{sec:sleight}

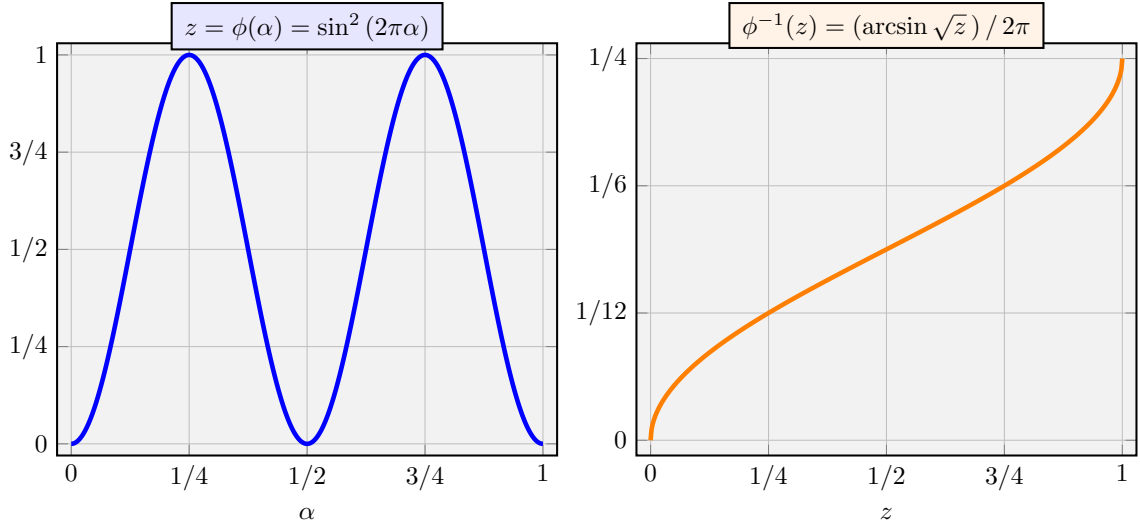
\begin{figure}
\centering
\begin{tikzpicture}[scale=0.96]
\begin{axis}[xmin=-0.03, xmax=1.03, ymin=-0.03, ymax=1.03, legend pos=north west, legend style={empty legend, fill=blue!10, at={(0.5,1.1)},anchor=north}, axis background/.style={fill=gray!10}, xtick={0, 0.25, 0.5, 0.75, 1}, xticklabels={0,1/4,1/2,3/4,1},
ytick={0, 0.25, 0.5, 0.75, 1}, yticklabels={0, 1/4, 1/2, 3/4, 1}, thick, minor x tick num=0, minor y tick num=0, grid=both, xlabel=$\alpha$]
\addplot[samples=500,domain=0:1, color=blue,line width=1.8pt]{ (sin(2*pi*deg(x)))^2 };
\addlegendentry{$z = \phi(\alpha) = \sin^2 \left( 2\pi \alpha \right)$};
\end{axis}
\end{tikzpicture}
\begin{tikzpicture}[scale=0.96]
\begin{axis}[xmin=-0.03, xmax=1.03, ymin=-0.01, ymax=0.26, legend pos=north west, legend style={empty legend, fill=orange!10, at={(0.5,1.1)},anchor=north}, axis background/.style={fill=gray!10}, xtick={0, 0.25, 0.5, 0.75, 1}, xticklabels={0, 1/4, 1/2, 3/4, 1},
ytick={0,0.08333333,0.1666667,0.25}, yticklabels={0,1/12,1/6,1/4}, thick, minor x tick num=0, minor y tick num=0, grid=both, xlabel=$z$]
\addplot[samples=500,domain=0:1, color=orange,line width=1.8pt]{rad(asin(sqrt(x))) / (2*pi)};
\addlegendentry{$\phi^{-1} (z) = \left( \arcsin{\sqrt{z}} \, \right) / \, 2\pi $};
\end{axis}
\end{tikzpicture}
\caption{Illustrations of the homeomorphism~$\phi$ and of its inverse~$\phi^{-1}$. Note that a continuous function with a continuous inverse function, as is the case here, is called a homeomorphism.}
\label{fig:Phi}  
\end{figure}

Let us assume that for a given~$k \in \mathbb{N}$ the value~$z_k$ of an iterate produced by the logistic map can be linked to some other variable~$\alpha_k$ through the following change of variable:
\begin{equation}
z_k \equiv \phi(\alpha_k) = \sin^2(2\pi \alpha_k)
\label{eq::definePhi}
\end{equation}
Obviously, we will soon try and identify~$\alpha_k$ as an iterate of the dyadic transformation~$\mathcal{D}$ defined in~eq.(\ref{eq:dyadic}).  For now, let us continue by following the evolution of~$z_k$ as prescribed by the logistic map.  Using the newly-introduced~$\phi$, the value of the next iterate~$z_{k+1} \equiv \mathcal{L}(z_k)$ can be expressed as:
\begin{equation*}
z_{k+1} \equiv 4 \phi(\alpha_k) \big( 1 - \phi(\alpha_k) \big) = 4 \sin^2(2\pi \alpha_k) \cos^2(2\pi \alpha_k) \quad \Longrightarrow \quad z_{k+1} = \sin^2 \left( 2\pi \, 2\alpha_k \right)
\end{equation*}
In order to make a connection, let us recognize that:
\begin{equation*}
\sin^2 \big( 2\pi \mathcal{D}(\alpha_k) \big) = 
\begin{cases}
\sin^2 \left( 2\pi 2\alpha_k \right) & (\alpha_k < 1/2) \\
\sin^2 \big( 2\pi (2\alpha_k  - 1)  \big) = \sin^2 \left( 2\pi 2\alpha_k - 2\pi \right) =  \sin^2 \left( 2\pi 2\alpha_k \right) & (\alpha_k > 1/2) \\
\end{cases}
\end{equation*}
due to the periodicity of~$\phi$ as one can see in~Figure~\ref{fig:Phi}.  Plugging this result back into the expression for~$z_{k+1}$ and introducing~$\alpha_{k+1} \equiv \mathcal{D}(\alpha_k)$ leads to the following identity:
\begin{equation*}
z_{k+1} = \phi(\alpha_{k+1})  \quad \Leftrightarrow \quad \mathcal{L}(z_k) = \phi \big( \mathcal{D}(\alpha_k) \big)
\label{eq:topologicalConjugacy}
\end{equation*}

\noindent This result is nothing short of astonishing: it proves that although~$\mathcal{L}$ and~$\mathcal{D}$ superficially look like very different recurrence relations, it is more appropriate to think of them as topological twins that share identical dynamics.  With~$\phi$ and its inverse~$\phi^{-1}$ acting as a bridges between the two topological spaces, this marvelous behavior known as topological conjugacy can be summarized pictorially via the commutative diagram:

\begin{center} \begin{tikzpicture}

\node[] (zks) at (0, 0) {};
\node[] (zk0) at (2, 0) {};
\node[] (zk) at (4, 0) {$z_k$ };
\node[] (zk1) at (7, 0) {$z_{k+1}$ };
\node[] (zk2) at (10, 0) {$z_{k+2}$ };
\node[] (zk3) at (12, 0) {};
\node[] (zkf) at (14, 0) {};

\node[] (tks) at (0, 2) {};
\node[] (tk0) at (2, 2) {};
\node[] (tk) at (4, 2) {$\alpha_k$ };
\node[] (tk1) at (7, 2) {$\alpha_{k+1}$ };
\node[] (tk2) at (10, 2) {$\alpha_{k+2}$ };
\node[] (tk3) at (12, 2) {};
\node[] (tkf) at (14, 2) {};

\draw[->] (zk.east) to (zk1.west) node[xshift=-1.0cm, yshift=0.0cm, above] {$\mathcal{L}$ };
\draw[->] (zk1.east) to (zk2.west) node[xshift=-1.0cm, yshift=0.0cm,above] {$\mathcal{L}$ };
\draw[dashed] (zk2.east) to (zk3.west) node[] {};
\draw[dashed] (zk0.east) to (zk.west) node[] {};

\draw[->] (tk.east) to (tk1.west) node[xshift=-1.0cm, yshift=0.0cm,below] {$\mathcal{D}$ };
\draw[->] (tk1.east) to (tk2.west) node[xshift=-1.0cm, yshift=0.0cm,below] {$\mathcal{D}$ };
\draw[dashed] (tk2.east) to (tk3.west) node[] {};
\draw[dashed] (tk0.east) to (tk.west) node[] {};

\draw[dashed] (tks.east) to (tk0.west) node[] {};
\draw[dashed] (zks.east) to (zk0.west) node[] {};

\draw[dashed] (zk3.east) to (zkf.west) node[] {};
\draw[dashed] (tk3.east) to (tkf.west) node[] {};

\draw[->, thick, orange] (zk0.west) to [bend left] (tk0.west) node[xshift=-0.3cm, yshift=-1.0cm,left] {$\phi^{-1}$ } ;
\draw[->, thick, blue] (tk0.east) to [bend left] (zk0.east) node[xshift=0.3cm, yshift=1.0cm,right] {$\phi$ } ;

\draw[->, thick, blue] (tk3.east) to [bend left] (zk3.east) node[xshift=0.3cm, yshift=1.0cm,right] {$\phi$ } ;
\draw[->, thick, orange] (zk3.west) to [bend left] (tk3.west) node[xshift=-0.3cm, yshift=-1.0cm,left] {$\phi^{-1}$ } ;

\end{tikzpicture} \end{center}

\noindent In other words, one may decide to study the dynamics of a variable according to the logistic map and never leave this representation but, all the while, think about how this dynamics is equivalently represented in some kind of ``parallel universe'' where variables evolve according to the dyadic transformation up to a compositional ``portal'' in the form of~$\phi$. \\

\noindent Now that we have this remarkable phenomenon in our hands, how can we exploit it to apply makeup to the basic encoding/decoding strategy presented in the previous section?

\subsection{Construction of the initial condition}

Instead of populating the initial condition with direct binary approximations to the components of the original dataset, let us start with a pre-processing step by applying~$\phi^{-1}$ to all of the samples of~$\mathcal{X}=[x_0,x_1,x_2,\cdots,x_n]$ before going into a fixed-point binary representation:

\begin{center} \begin{tikzpicture}
\node[] (x0) at (0, 0) {$x_0$ };
\node[] (x1) at (0, -1) {$x_1$ };
\node[] (xdots) at (0, -2) {$\vdots$ };
\node[] (xn) at (0, -3) {$x_n$ };

\node[] (y0) at (2, 0) {$\phi^{-1} (x_0) \approx $ };
\node[] (y1) at (2, -1) {$\phi^{-1} (x_1) \approx $ };
\node[] (yDots) at (2, -2) {$\vdots$ };
\node[] (yn) at (2, -3) {$\phi^{-1} (x_n) \approx $ };

\node[myboxFst] (a1) at (3.3, 0) {  };
\node[myboxFst] (a2) at (3.8, 0) { };
\node[myboxFst] (a3) at (4.3, 0) {  };
\node[myboxFst] (a4) at (4.8, 0) { };
\node[myboxFst] (a5) at (5.3, 0) {  };
\node[myboxFst] (a6) at (5.8, 0) { };
\node[myboxFst] (a7) at (6.3, 0) { };
\node[myboxFst] (a8) at (6.8, 0) { };

\node[myboxSec] (b1) at (3.3, -1) {  };
\node[myboxSec] (b2) at (3.8, -1) { };
\node[myboxSec] (b3) at (4.3, -1) {  };
\node[myboxSec] (b4) at (4.8, -1) { };
\node[myboxSec] (b5) at (5.3, -1) {  };
\node[myboxSec] (b6) at (5.8, -1) { };
\node[myboxSec] (b7) at (6.3, -1) { };
\node[myboxSec] (b8) at (6.8, -1) { };

\node[myboxLst] (c1) at (3.3, -3) {  };
\node[myboxLst] (c2) at (3.8, -3) { };
\node[myboxLst] (c3) at (4.3, -3) {  };
\node[myboxLst] (c4) at (4.8, -3) { };
\node[myboxLst] (c5) at (5.3, -3) {  };
\node[myboxLst] (c6) at (5.8, -3) { };
\node[myboxLst] (c7) at (6.3, -3) { };
\node[myboxLst] (c8) at (6.8, -3) { };

\draw[->] (x0.east) to (y0.west) node[ ] { };
\draw[->] (x1.east) to (y1.west) node[ ] { };
\draw[->] (xn.east) to (yn.west) node[ ] { };

\draw [decorate,decoration={brace,amplitude=10pt}]
(a8.north east) -- (c8.south east) node [align=left, black,midway,xshift=4.3cm] 
{Convert all samples from the original data space \\ into fixed-point binary representation with~$\tau$ bits \\ of precision in the conjugate (dyadic) space};

\end{tikzpicture}
\end{center}

\noindent Concatenating all these transformed data points into a long binary string composed of~$\sim n\tau$ bits associated with weight factors of decreasing significance factor (see section~\ref{naive::init}) leads to the following initial condition: \\

\noindent \makebox[\textwidth][c]{\begin{tikzpicture}   

\node[] (alpha) at (0.2, 0) {$\alpha_{0} = $ };

\node[myboxFst] (a1) at (1, 0) { \phantom{0} };
\node[myboxFst] (a2) at (1.5, 0) { \phantom{0}  };
\node[myboxFst] (a3) at (2, 0) {\phantom{0} };
\node[myboxFst] (a4) at (2.5, 0) {\phantom{0} };
\node[myboxFst] (a5) at (3, 0) { \phantom{0} };
\node[myboxFst] (a6) at (3.5, 0) { \phantom{0} };
\node[myboxFst] (a7) at (4, 0) {\phantom{0} };
\node[myboxFst] (a8) at (4.5, 0) {\phantom{0} };
\node[myboxSec] (b1) at (5, 0) { \phantom{0} };
\node[myboxSec] (b2) at (5.5, 0) { \phantom{0}};
\node[myboxSec] (b3) at (6, 0) {\phantom{0}  };
\node[myboxSec] (b4) at (6.5, 0) {\phantom{0} };
\node[myboxSec] (b5) at (7, 0) {\phantom{0}  };
\node[myboxSec] (b6) at (7.5, 0) {\phantom{0} };
\node[myboxSec] (b7) at (8, 0) {\phantom{0} };
\node[myboxSec] (b8) at (8.5, 0) {\phantom{0} };

\node[myboxLst] (l1) at (9, 0) {\phantom{0} };
\node[myboxLst] (l2) at (9.5, 0) {\phantom{0} };
\node[myboxLst] (l3) at (10, 0) {\phantom{0} };
\node[myboxLst] (l4) at (10.5, 0) {\phantom{0} };
\node[myboxLst] (l5) at (11, 0) {\phantom{0} };
\node[myboxLst] (l6) at (11.5, 0) {\phantom{0} };
\node[myboxLst] (l7) at (12, 0) {\phantom{0} };
\node[myboxLst] (l8) at (12.5, 0) {\phantom{0} };

\node[minimum width=0.5cm] (bDots) at (13.15, 0) {  $\cdots$ };
\node[minimum width=0.5cm] (bDots2) at (13.65, 0) {  $\cdots$ };
\node[minimum width=0.5cm] (bDots3) at (14.15, 0) {  $\cdots$ };

\node[myboxLstFinal] (y1) at (14.8, 0) {\phantom{0} };
\node[myboxLstFinal] (y2) at (15.3, 0) {\phantom{0} };
\node[myboxLstFinal] (y3) at (15.8, 0) {\phantom{0} };
\node[myboxLstFinal] (y4) at (16.3, 0) {\phantom{0} };
\node[myboxLstFinal] (y5) at (16.8, 0) {\phantom{0} };
\node[myboxLstFinal] (y6) at (17.3, 0) {\phantom{0} };
\node[myboxLstFinal] (y7) at (17.8, 0) {\phantom{0} };
\node[myboxLstFinal] (y8) at (18.3, 0) {\phantom{0} };

\node[myboxFst, fill=gray!10] (basis1) at (1, -1) {$1/2$};
\node[myboxFst, fill=gray!10] (basis2) at (4.5, -1) {$1/2^\tau$};
\node[myboxSec, fill=gray!10] (basis3) at (8.5, -1) {$1/2^{2\tau}$ };
\node[myboxSec, fill=gray!10] (basis4) at (12.5, -1) {$1/2^{3\tau}$ };
\node[myboxSec, fill=gray!10] (basis5) at (14.8, -1) {$1/2^{n\tau}$ };
\node[myboxSec, fill=gray!10] (basis6) at (18.3, -1) {$1/2^{(n+1)\tau}$ };

\draw [decorate,decoration={brace,amplitude=10pt},xshift=60pt]
(y1.north west) -- (y8.north east) node [black,midway,yshift=0.3cm,above] 
{$\approx \phi^{-1}(x_n)  / 2^{n\tau}$};

\draw[->, dashed] (a1.south) to (basis1.north) node[ ] { };
\draw[->, dashed] (a8.south) to (basis2.north) node[ ] { };
\draw[->, dashed] (b8.south) to (basis3.north) node[ ] { };
\draw[->, dashed] (l8.south) to (basis4.north) node[ ] { };
\draw[->, dashed] (y1.south) to (basis5.north) node[ ] { };
\draw[->, dashed] (y8.south) to (basis6.north) node[ ] { };

\draw [decorate,decoration={brace,amplitude=10pt},xshift=60pt]
(a1.north west) -- (a8.north east) node [black,midway,yshift=0.3cm,above] 
{$\approx \phi^{-1}(x_0)$};

\draw [decorate,decoration={brace,amplitude=10pt},xshift=60pt]
(b1.north west) -- (b8.north east) node [black,midway,yshift=0.3cm,above] 
{$\approx \phi^{-1}(x_1)  / 2^{\tau} $};

\draw [decorate,decoration={brace,amplitude=10pt},xshift=60pt]
(l1.north west) -- (l8.north east) node [black,midway,yshift=0.3cm,above] 
{$\approx \phi^{-1}(x_2)   / 2^{2\tau} $};

\draw [decorate,decoration={mirror,brace,amplitude=10pt},xshift=60pt] (basis1.south) -- (basis2.south) node [black,midway,yshift=-0.3cm,below] {$\alpha_0 \approx \phi^{-1}(x_0)$};

\end{tikzpicture}} \\

\noindent which we immediately convert to its corresponding decimal form via~$\alpha_0 = \sum_{i=1}^{(n+1)\tau} \alpha_i / 2^i$ keeping in mind the required number of digits of precision (see section~\ref{sec::decimalPrecision}).  Having built~$\alpha_0$ in the conjugate (dyadic) space, it is now time to use topological conjugacy in order to transport this initial condition back out into the original data domain by applying the~$\phi$ ``portal''.  In other words, we consider an initial condition~$z_0$ defined as:
\begin{equation*}
\highlight[black!10]{z_0 = \phi(\alpha_0) =  \sin^2 \left( 2 \pi \alpha_0 \right)}
\end{equation*}

\noindent Because of our careful construction of the first~$\tau$ significant bits leading to~$\alpha_0 \approx \phi^{-1} (x_0)$ and the fact that~$\phi$ is a homeomorphism, we see that~$z_0 = \phi(\alpha_0) \approx x_0$ is a good approximation to the first sample:
\begin{equation*}
\highlight[black!10]{z_0 \approx x_0 \,\,\,\, \text{with error bound} \,\, \left| z_0 - x_0 \right| < \pi / 2^{\tau-1}}
\end{equation*}
The error bound between the generated value~$z_0$ and the true value of the first sample~$x_0$ of~$\mathcal{X}$ will be explained later in section~\ref{ref::lipschitz} but is already reported here for the sake of completeness.

\subsection{Decoding using the logistic map}

Thanks to the topological conjugacy relationship between the dyadic transformation and the logistic map, we can plug in the solution~$\alpha_k \equiv \mathcal{D}^{k\tau}(\alpha_0) =  2^{k\tau}\alpha_0 \, \text{mod} \, 1$ obtained previously in the dyadic space, see~eq.(\ref{eq:exactSol}), into its conjugate counterpart~$\highlight[black!10]{z_k \equiv \mathcal{L}^{k\tau} (z_0)} = \phi(\alpha_k) = \phi(2^{k\tau} \alpha_0 )$.  Note that because the codomain of~$\phi$ is restricted to the unit interval, the modulo operation initially present in~$\alpha_k$ is not relevant anymore for~$z_k$.  Finally, we relate the initial condition~$\alpha_0$ constructed above (converted to decimal representation) to its conjugate value~$z_0$ via~$\alpha_0 = \phi^{-1} (z_0) = ( \arcsin{\sqrt{z_0}} ) /2\pi$.  All together, the iterated evolution of the logistic map admits an exact solution \dSmiley \dSmiley \dSmiley \underline{\bf functionally identical to~$f_\alpha$} \dSmiley \dSmiley \dSmiley our promised crown jewel that was prophesied on the first page of this paper.  Continuing with the notation used here, this decoding function is given by:
\begin{equation}
\highlight[black!10]{z_k = \sin^2 \left( 2^{k\tau} \arcsin{\sqrt{z_0}}  \right)}
\label{eq:logisticExact}
\end{equation}

\noindent We have already seen that the first sample is well approximated with~$z_0 \approx x_0$.  Because of topological conjugacy, we will now show that all other samples can also be recovered by sampling~$z_k$ at~$k\in[0,\cdots,n]$. \\

\noindent To gain more insight into why the dynamics of the logistic map reproduces the remaining samples of~$\mathcal{X}$, let us consider the first iteration of~$z_0$ under~$\mathcal{L}$.  Following the ``parallel universe'' interpretation of topological conjugacy depicted in the diagram of paragraph~\ref{sec:sleight}, we see that the first iterate~$z_1 = \mathcal{L}^\tau(z_0)$ is conjugate to~$\alpha_1 = \mathcal{D}^\tau(\alpha_0)$.  In this conjugate (dyadic) space, the combined effect of~$\tau$ successive iterations of~$\mathcal{D}$ results in a leftwise block shift of the first~$\tau$ bits of the initial condition~$\alpha_0$.  Effectively, the bit sequence that was originally encoding~$\phi^{-1}(x_0)$ is discarded (moved over the binary point and turned to~0 by the modulo operation) making room for the next sequence of~$\tau$ bits encoding~$\phi^{-1}(x_1)$ to become the leading contribution to the numerical value of~$\alpha_1$.    \\ \\
\noindent \makebox[\textwidth][c]{\begin{tikzpicture}   

\node[] (alpha) at (-0.6, 0) {$\alpha_1 = \mathcal{D}^\tau \left( \alpha_0 \right) = $ };

\node[myboxSec] (a1) at (1, 0) { \phantom{0} };
\node[myboxSec] (a2) at (1.5, 0) { \phantom{0}  };
\node[myboxSec] (a3) at (2, 0) {\phantom{0} };
\node[myboxSec] (a4) at (2.5, 0) {\phantom{0} };
\node[myboxSec] (a5) at (3, 0) { \phantom{0} };
\node[myboxSec] (a6) at (3.5, 0) { \phantom{0} };
\node[myboxSec] (a7) at (4, 0) {\phantom{0} };
\node[myboxSec] (a8) at (4.5, 0) {\phantom{0} };

\node[myboxLst] (b1) at (5, 0) { \phantom{0} };
\node[myboxLst] (b2) at (5.5, 0) { \phantom{0}};
\node[myboxLst] (b3) at (6, 0) {\phantom{0}  };
\node[myboxLst] (b4) at (6.5, 0) {\phantom{0} };
\node[myboxLst] (b5) at (7, 0) {\phantom{0}  };
\node[myboxLst] (b6) at (7.5, 0) {\phantom{0} };
\node[myboxLst] (b7) at (8, 0) {\phantom{0} };
\node[myboxLst] (b8) at (8.5, 0) {\phantom{0} };

\node[minimum width=0.5cm] () at (9.15, 0) {  $\cdots$ };
\node[minimum width=0.5cm] () at (9.65, 0) {  $\cdots$ };
\node[minimum width=0.5cm] () at (10.15, 0) {  $\cdots$ };

\node[myboxLstFinal] (y1) at (10.8, 0) {\phantom{0} };
\node[myboxLstFinal] (y2) at (11.3, 0) {\phantom{0} };
\node[myboxLstFinal] (y3) at (11.8, 0) {\phantom{0} };
\node[myboxLstFinal] (y4) at (12.3, 0) {\phantom{0} };
\node[myboxLstFinal] (y5) at (12.8, 0) {\phantom{0} };
\node[myboxLstFinal] (y6) at (13.3, 0) {\phantom{0} };
\node[myboxLstFinal] (y7) at (13.8, 0) {\phantom{0} };
\node[myboxLstFinal] (y8) at (14.3, 0) {\phantom{0} };

\node[myboxFst, fill=gray!10] (basis1) at (1, -1) {$1/2$};
\node[myboxFst, fill=gray!10] (basis2) at (4.5, -1) {$1/2^\tau$};
\node[myboxSec, fill=gray!10] (basis3) at (8.5, -1) {$1/2^{2\tau}$ };
\node[myboxSec, fill=gray!10] (basis5) at (10.8, -1) {$1/2^{(n-1)\tau}$ };
\node[myboxSec, fill=gray!10] (basis6) at (14.3, -1) {$1/2^{n\tau}$ };

\draw[->, dashed] (a1.south) to (basis1.north) node[ ] { };
\draw[->, dashed] (a8.south) to (basis2.north) node[ ] { };
\draw[->, dashed] (b8.south) to (basis3.north) node[ ] { };
\draw[->, dashed] (y1.south) to (basis5.north) node[ ] { };
\draw[->, dashed] (y8.south) to (basis6.north) node[ ] { };

\draw [decorate,decoration={brace,amplitude=10pt},xshift=60pt]
(a1.north west) -- (a8.north east) node [black,midway,yshift=0.3cm,above] 
{$\approx \phi^{-1}(x_1)$};

\draw [decorate,decoration={brace,amplitude=10pt},xshift=60pt]
(b1.north west) -- (b8.north east) node [black,midway,yshift=0.3cm,above] 
{$\approx \phi^{-1}(x_2) / 2^\tau$};

\draw [decorate,decoration={brace,amplitude=10pt},xshift=60pt]
(y1.north west) -- (y8.north east) node [black,midway,yshift=0.3cm,above] 
{$\approx \phi^{-1}(x_n)  / 2^{(n-1)\tau}$};

\draw [decorate,decoration={mirror,brace,amplitude=10pt},xshift=60pt] (basis1.south) -- (basis2.south) node [black,midway,yshift=-0.3cm,below] {$\alpha_1 \approx \phi^{-1}(x_1)$};

\end{tikzpicture}} \\

\noindent Thanks to our portal between topological spaces, the data space sample~$z_1$ is conjugate to~$\alpha_1 \approx \phi^{-1}(x_1)$ meaning that~$z_1 = \phi(\alpha_1) \approx x_1$.  This result confirms that appying~eq.(\ref{eq:logisticExact}) with~$k=1$ leads to:
\begin{equation*}
\highlight[black!10]{z_1 = \sin^2 \left( 2^\tau \arcsin{\sqrt{z_0}}  \right)  \approx x_1 \,\,\,\, \text{with error bound} \,\,\,\, \left| z_1 - x_1 \right| < \pi / 2^{\tau-1}}    
\end{equation*}

\noindent The error bound will be explained in the next paragraph.  Moving on to the second iteration, $z_2 = \mathcal{L}^{2\tau}(z_0)$ is conjugate to~$\alpha_2 = \mathcal{D}^{2\tau}(\alpha_0)$ which is represented in the conjugate space by: \\ \\
\noindent \begin{tikzpicture}   

\node[] (alpha) at (-0.7, 0) {$\alpha_{2} = \mathcal{D}^{2\tau} \left( \alpha_0 \right) = $ };

\node[myboxLst] (a1) at (1, 0) { \phantom{0} };
\node[myboxLst] (a2) at (1.5, 0) { \phantom{0}  };
\node[myboxLst] (a3) at (2, 0) {\phantom{0} };
\node[myboxLst] (a4) at (2.5, 0) {\phantom{0} };
\node[myboxLst] (a5) at (3, 0) { \phantom{0} };
\node[myboxLst] (a6) at (3.5, 0) { \phantom{0} };
\node[myboxLst] (a7) at (4, 0) {\phantom{0} };
\node[myboxLst] (a8) at (4.5, 0) {\phantom{0} };

\node[minimum width=0.5cm] () at (5.15, 0) {  $\cdots$ };
\node[minimum width=0.5cm] () at (5.65, 0) {  $\cdots$ };
\node[minimum width=0.5cm] () at (6.15, 0) {  $\cdots$ };

\node[myboxLstFinal] (y1) at (6.8, 0) {\phantom{0} };
\node[myboxLstFinal] (y2) at (7.3, 0) {\phantom{0} };
\node[myboxLstFinal] (y3) at (7.8, 0) {\phantom{0} };
\node[myboxLstFinal] (y4) at (8.3, 0) {\phantom{0} };
\node[myboxLstFinal] (y5) at (8.8, 0) {\phantom{0} };
\node[myboxLstFinal] (y6) at (9.3, 0) {\phantom{0} };
\node[myboxLstFinal] (y7) at (9.8, 0) {\phantom{0} };
\node[myboxLstFinal] (y8) at (10.3, 0) {\phantom{0} };

\node[myboxFst, fill=gray!10] (basis1) at (1, -1) {$1/2$};
\node[myboxFst, fill=gray!10] (basis2) at (4.5, -1) {$1/2^\tau$};
\node[myboxSec, fill=gray!10] (basis5) at (6.8, -1) {$1/2^{(n-2)\tau}$ };
\node[myboxSec, fill=gray!10] (basis6) at (10.3, -1) {$1/2^{(n-1)\tau}$ };

\draw [decorate,decoration={brace,amplitude=10pt},xshift=60pt] (y1.north west) -- (y8.north east) node [black,midway,yshift=0.3cm,above] {$\approx \phi^{-1}( x^n) / 2^{(n-2)\tau}  $};

\draw [decorate,decoration={brace,amplitude=10pt},xshift=60pt] (a1.north west) -- (a8.north east) node [black,midway,yshift=0.3cm,above] {$\approx \phi^{-1}(x^2)$};

\draw [decorate,decoration={mirror,brace,amplitude=10pt},xshift=60pt] (basis1.south) -- (basis2.south) node [black,midway,yshift=-0.3cm,below] {$\alpha_{2} \approx \phi^{-1}( x^2)$};

\draw[->, dashed] (a1.south) to (basis1.north) node[ ] { };
\draw[->, dashed] (a8.south) to (basis2.north) node[ ] { };
\draw[->, dashed] (y1.south) to (basis5.north) node[ ] { };
\draw[->, dashed] (y8.south) to (basis6.north) node[ ] { };

\end{tikzpicture} \\ 

\noindent Once again, invoking topological conjugacy arguments to the observation that~$\alpha_2 \approx \phi^{-1}(x_2)$ confirms that the third sample is indeed recovered~$z_2 = \phi(\alpha_2) \approx x_2$ by sampling~eq.(\ref{eq:logisticExact}) at~$k=2$.  In other words:
\begin{equation*}
\highlight[black!10]{z_2 = \sin^2 \left( 2^{2\tau} \arcsin{\sqrt{z_0}}  \right)  \approx x_2 \,\,\,\, \text{with error bound} \,\,\,\, \left| z_2 - x_2 \right| < \pi / 2^{\tau-1}}  
\end{equation*}

\noindent Pushing this logic~$n$ times, it is clear that the binary representation of~$\alpha_0$ in the conjugate space gets gradually stripped off of all of its leading components finally leaving only the bits corresponding to the last transformed data point of~$\mathcal{X}$.  In picture: \\ \\
\noindent \begin{tikzpicture}   

\node[] (alpha) at (-0.8, 0) {$\alpha_n = \mathcal{D}^{n\tau} \left( \alpha_0 \right) = $ };

\node[myboxLstFinal] (a1) at (1, 0) { \phantom{0} };
\node[myboxLstFinal] (a2) at (1.5, 0) { \phantom{0}  };
\node[myboxLstFinal] (a3) at (2, 0) {\phantom{0} };
\node[myboxLstFinal] (a4) at (2.5, 0) {\phantom{0} };
\node[myboxLstFinal] (a5) at (3, 0) { \phantom{0} };
\node[myboxLstFinal] (a6) at (3.5, 0) { \phantom{0} };
\node[myboxLstFinal] (a7) at (4, 0) {\phantom{0} };
\node[myboxLstFinal] (a8) at (4.5, 0) {\phantom{0} };

\node[myboxFst, fill=gray!10] (basis1) at (1, -1) {$1/2$};
\node[myboxFst, fill=gray!10] (basis2) at (4.5, -1) {$1/2^\tau$};

\draw [decorate,decoration={brace,amplitude=10pt},xshift=60pt] (a1.north west) -- (a8.north east) node [black,midway,yshift=0.3cm,above] {$\phi^{-1}(x_n)$};

\draw[->, dashed] (a1.south) to (basis1.north) node[ ] { };
\draw[->, dashed] (a8.south) to (basis2.north) node[ ] { };

\end{tikzpicture} \\

\noindent confirming that~$\phi^{-1}(x_n)$, originally constructed as the least signifiant~$\tau$ bits of the initial condition~$\alpha_0$, becomes the leading contribution to~$\alpha_n$.  Therefore~$z_n = \phi(\alpha_n) \approx x_n$ is correctly decoded:
\begin{equation*}
\highlight[black!10]{z_n = \sin^2 \left( 2^{n\tau} \arcsin{\sqrt{z_0}}  \right)  \approx x_n \,\,\,\, \text{with error bound} \,\,\,\, \left| z_n - x_n \right| < \pi / 2^{\tau-1}}  
\end{equation*}

\noindent Before moving on to a detailed explanation of the error bound, let us stress how far we have gone: the entire dataset~$\mathcal{X}$ has been reproduced by the beautiful~$f_\alpha$ as promised in the introduction!

\subsection{Error bound: Lipschitz continuity}
\label{ref::lipschitz}

In order to complete the proof, let us justify the error bound~$\left| z_j - x_j \right| < \pi / 2^{\tau-1}$ between a sample~$x_j$ drawn from the dataset~$\mathcal{X}$ and its corresponding approximation~$z_j = \sin^2 \left( 2^{j\tau} \arcsin{\sqrt{z_0}} \right) \approx x_j$ given by~eq.(\ref{eq:logisticExact}) (which is identical to~$f_\alpha$ shown in the introduction). \\

\noindent According to the \href{https://en.wikipedia.org/wiki/Mean\_value\_theorem}{mean value theorem}, given a function~$\phi$ differentiable over an interval~$(a, b)$, there exists a point~$\xi \in (a,b)$ in this interval such that~$\phi^\prime(\xi) = \big( \phi(a) - \phi(b) \big) / \big(a-b \big)$.  Separately, it is easy to verify that the homeomorphism~$\phi$, our portal between different topological spaces defined in~eq.(\ref{eq::definePhi}), is \href{https://en.wikipedia.org/wiki/Lipschitz\_continuity}{Lipschitz continuous}: its derivative is bounded by~$\left| \phi^\prime(\xi) \right| = 2 \pi \left| \sin 4\pi \alpha \right| \leq 2 \pi$.  Plugging this Lipschitz constant of~$2\pi$ into the expression of the mean value theorem leads to:
\begin{equation*}
\left| \phi(a) - \phi(b) \right| \leq  2\pi \left| a - b \right|
\end{equation*}

\noindent We can now specialize this result to the case where~$a = \alpha_j$ and~$b = \phi^{-1}(x_j)$.  Thanks to the topological conjugacy relationship~$z_j = \phi(\alpha_j)$ and the inequality above, we get:
\begin{equation*}
\left| z_j - x_j  \right|  \leq 2\pi \left| \alpha_j - \phi^{-1}(x_j) \right|
\end{equation*}
Because~$\phi^{-1}(x_j)$ weighs in as the leading contribution to the numerical value of~$\alpha_j$ (up to its first~$\tau$ bits), we have~$\alpha_j \approx \phi^{-1}(x_j)$ with an error bound~$\left| \alpha_j - \phi^{-1}(x_j) \right| < 1/2^\tau$ (see section~\ref{lab:naive}).  Combining this inequality together with the previous one derived above leads to the promised error bound:
\begin{equation*}
\highlight[black!10]{\left| z_j - x_j  \right| < \dfrac{\pi}{2^{\tau-1}} }
\end{equation*}

\noindent This shows that using the logistic map to decode the dataset generates a degradation by a factor of~$2\pi$ compared to what can be achieved with the more straightforward strategy described in section~\ref{lab:naive}.

\subsection{Summary in code}

Inspired by the previous section, we conclude by showing how the algorithm can easily be implemented in just a few lines of code.  {\it (Helper functions to convert between binary and decimal representations, using a default value of~$\tau=8$, as well as the construction of a dataset~$\mathcal{X}$ composed of~50 random numbers in the unit interval are provided in appendix~\ref{code::appendix}.)}  \\

\noindent We start by applying~$\phi^{-1}$ (denoted by~\texttt{phiInv}) to all the samples of~$\mathcal{X}$ and follow by expressing the resulting values as fixed-point precision (defaulted to~$\tau$ bits) binary numbers.  Concatenating all these transformed samples together yields the conjugate initial value~$\alpha_0$ in binary.  \\
\begin{lstlisting}[language=Python, caption=]
# xs: List[float] is the dataset instantiated here as a list of random samples with decimal
#     values defined the Appendix

phiInv                 = lambda z: np.arcsin(np.sqrt(z)) / (2 * np.pi)
decimalToBinary_phiInv = lambda z: decimalToBinary(phiInv(z))

conjugateInitial_binary = ''.join(map(decimalToBinary_phiInv, xs))
print('conjugateInitial_binary = %s\n' % conjugateInitial_binary)

conjugateInitial_binary = 001000010010100100100100001000010001110000100110000111010011001000
                          111000000110110010110000100001001000100011010000001010000011000000
                          010100101110001011000011000000111010001011010001111000101100000011
                          100010010100001111001101100010000000011100000101100010101100011110
                          001000100000010100100100001001000010010000110110001001110001101000
                          011101001010000000101000100110001001110001001100001110000110000001
                          1010

conjugateInitial = binaryToDecimal(conjugateInitial_binary)
print('conjugateInitial = %s' % conjugateInitial)

conjugateInitial = 0.12953401382778691458786695916416542476624903080900276738757903052119237
                    703845826610929029135345417865945367623690935237

\end{lstlisting}

\vspace{0.1cm}

\noindent As before, the~50 samples require~$50\times  (\tau=8)=400$ bits of precision which translates to~$\approx 120$ digits of decimal precision as can be seen respectively in~\texttt{conjugateInitial\_binary} and~\texttt{conjugateInitial}.   \\

\noindent One peculiarity of having used the~$\phi^{-1}$ function is that every~$\tau^\text{th}$ bit of~\texttt{conjugateInitial\_binary} is~0.  This is because~$\phi^{-1}(z) < 1/4 \,\, , \,\, \forall z \in [0,1]$ as can be gleaned from Figure~\ref{fig:Phi}.  Since leading bits are associated with a weight~$1/2$, the first non-zero bit for real numbers~$<1/4$ is always the second bit.  Although this introduces a detectable redundancy which causes additional~$\approx 10\%$ memory requirements, more efficient strategies come at the cost of extra complexity that we do not wish to explore here. \\

\noindent The next step consists in transporting the decimal version of the initial condition~$\alpha_0 \equiv$~\texttt{conjugateInitial} from the conjugate space to the original data space by using the~$\phi$ transformation: \\
\begin{lstlisting}[language=Python, caption=]
from mpmath import pi as Pi
from mpmath import sin as Sin

# Note that (...) ** 2 is also overloaded and dispatches to mpmath.power(..., 2)

phi = lambda alpha: Sin(2 * Pi * alpha) ** 2

decimalInitial = phi(conjugateInitial)
print('decimalInitial = %s' % decimalInitial)

decimalInitial = 0.5284726382230582232141477613114233413442412684154899609425789100214256216
                 5617954914071030171163637703155000963962531642

\end{lstlisting}

\vspace{0.3cm}

\noindent As discussed in the code summary of the previous section, all arithmetic operations on~$z_0 \equiv$~\texttt{decimalInitial} are overloaded and dispatched to an external library that handles arbitrary-precision operations.  Finally, we implement the decoding function~\texttt{logisticDecoder} equivalent to the original~$f_\alpha$ that was obtained in~eq.(\ref{eq:logisticExact}). Sampling at~$k=[0,\cdots,n]$ builds a list~\texttt{decodedValues} that approximates the original dataset~$\mathcal{X}$. \\
\begin{lstlisting}[language=Python, caption=]
from mpmath import asin as Asin
from mpmath import sqrt as Sqrt

logisticDecoder = lambda k: Sin(2 ** (k * tau) * Asin(Sqrt(decimalInitial))) ** 2
decodedValues = [float(logisticDecoder(_)) for _ in range(len(xs))]
\end{lstlisting}

\vspace{0.1cm}

\noindent Finally, one may verify that the theoretical error bound of~$\pi / 2^{\tau -1}$ is indeed satisfied by all the datapoints by making sure that~\texttt{normalizedErrors} always stays below unity.

\vspace{0.1cm}

\begin{lstlisting}[language=Python, caption=]
maxError = Pi / 2 ** (tau - 1)

normalizedErrors = [abs(decodedValue - dataPoint) / maxError 
                    for decodedValue, dataPoint in zip(decodedValues, xs)]
\end{lstlisting}

\begin{figure}[hb!]
\centering
\begin{subfigure}{.48\textwidth}
  \centering
  \includegraphics[width=\linewidth]{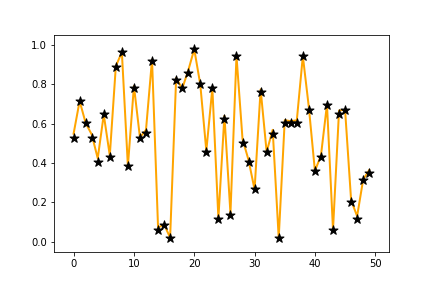}
\end{subfigure}%
\begin{subfigure}{.48\textwidth}
  \centering
  \includegraphics[width=\linewidth]{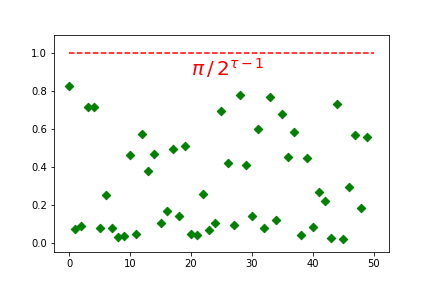}
\end{subfigure}
\caption{{\bf Left)} Comparison between the decoded values (black star markers) against the original dataset~$\mathcal{X}$ (thick orange line). {\bf Right)} Normalized error showing that the gap between decoded values and the original values never strays more than~$\pi/2^{\tau - 1}$ away (represented as the red horizontal dashed line) confirming the error bound derived in the main text. Notice that the error bound is now~$2\pi$ times larger than the one obtained using the more ``na\"ive'' encoding/decoding scheme of section~\ref{lab:naive}.}
\label{fig:advancedbound}
\end{figure}

\section{Data science perspective: what about generalization?}
\label{sec::generalization}

How to think about our wonderful~$f_\alpha(x) = \sin^2 \left( 2^{x\tau} \arcsin{\sqrt{\alpha}} \right)$ from the perspective of machine learning?  This simple and well-behaved (continuous, differentiable...) function can be ``trained'' to reproduce any dataset~$\mathcal{X}_\text{train}$ (up to an arbitrary level of precision determined by~$\tau \in \mathbb{N}$) by appropriately adjusting a single parameter~$\alpha \in \mathbb{R}$ ``learned'' from the training data. As a data scientist, an instinctive follow-up question would touch upon the generalization capabilities such a model may have on a previously unseen test dataset~$\mathcal{X}_\text{test}$.  \\

\begin{figure}[hb!]
\centering
\captionsetup{singlelinecheck=off}
\includegraphics[width=0.49\linewidth]{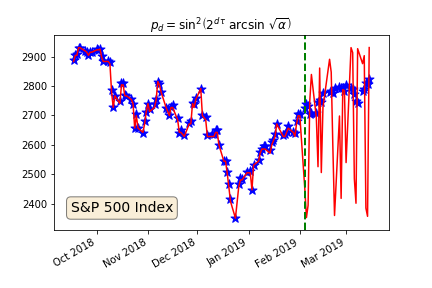}
\caption[]{Comparison between the daily price values of the S\&P 500 Index (blue markers) against the values predicted by~$p_d = \sin^2 \left( 2^{d\tau} \arcsin{\sqrt{\alpha}} \right)$ parametrized by:
\begin{align*}
\alpha = 0.&\big(\highlight[ao!10]{9186525008673170697061215177743819472103574383504939864690954692792184358812098296063847} \\
& \highlight[ao!10]{31739470802166549191011747211905687147014341039869287275246189278502982951415770973892328} \\
& \highlight[ao!10]{8994766865216570536672099485574178884250989741343121}\big) \,\, \begin{tikzpicture}[background rectangle/.style={fill=purple!10}, show background rectangle]
\draw[decorate sep={0.5mm}{2.5mm}, fill] (0,0) -- (5.7,0);
\end{tikzpicture} 
\end{align*}
where~$\approx 230$ decimal digits~(equivalent to~$\approx 760$ significant bits in binary) encoding the first~95 daily price values ($95 = 760 / \tau$ with default fixed-point precision~$\tau=8$ for each sample) are all explicitly provided.  If one continues to apply~$p_d$ beyond the information contained in the significant digits of~$\alpha$, as delimited by the vertical dashed green line, the predictions start to take on random values within the range seen in the ``training'' phase.
}
\label{fig:generalization}
\end{figure}

\noindent This is where the similarity with machine learning ends...  As should be clear by now, the parameter~$\alpha$ is not really learned from the training data as much as it is explicitly built as an alternative (encoded) representation of~$\mathcal{X}_\text{train}$ itself.  In  other words, the information content of~$\alpha$ is identical (i.e.~no compression besides a tunable recovery accuracy determined by~$\tau$) to that of~$\mathcal{X}_\text{train}$ and the model serves as nothing more than a mechanistic decoder.  Building on the previous visualizations,~$\alpha$ should be thought of pictorially as:
\begin{center}
\begin{tikzpicture}   
\node[] (alpha) at (-0.8, 0) {$\alpha = 0. $ };
\node[thick, draw=ao, fill=ao!10, ellipse, align=center] (a1) at (0.6, 0) { $\mathcal{X}_\text{train}$ };
\node[thick, draw=purple, fill=purple!10, ellipse, align=center] (a1) at (2.8, 0) { $\cdots \cdots \cdots$ };
\end{tikzpicture}
\end{center}
\noindent where the encoded version of the training data is followed by an (infinite) series of random digits.  In a way, decoding~$\mathcal{X}_\text{train}$ can be seen as probing~$\alpha$ at increasing levels of precision.  Once all the components of~$\mathcal{X}_\text{train}$ (green ellipsis above) have been extracted, we are starting to look at~$\alpha$ at a level of precision for which the digits are now random (red ellipsis above).  Despite providing an excellent fit to~$\mathcal{X}_\text{train}$, there is no reason to expect~$f_\alpha$ to provide any kind of generalization to data outside of~$\mathcal{X}_\text{train}$ as demonstrated in Figure~\ref{fig:generalization}. \\

\noindent  As already hinted at in the introduction,~$f_\alpha$ puts in perspective the common practice of measuring the complexity of a machine learning model by counting its total number of parameters and suggests that other techniques based on data compression (such as minimum description length...) may be more appropriate.  In addition, let us mention that given the huge number of parameters (each encoded as a floating-point number) available in modern deep learning architectures and the ease with which those models are able to fit randomly labeled data~\cite{intriguing,shamtikov}, one falls back to a timely and often asked question in machine learning research: How much can the values of the weights of a well-trained neural network be attributed to brute-force memorization of the training data vs. meaningful learning and why do they provide any kind of generalization?

\begin{appendices}

\section{Helper functions}
\label{code::appendix}

The ``summary in code'' paragraphs of sections~\ref{lab:naive} and~\ref{secondStrategy} invoke a number of helper functions needed to perform common tasks such as~i)~converting between decimal/binary representations and~ii)~generating a playground dataset~$\mathcal{X}$ composed of random values. All code fragments throughout the paper are written in the \href{https://www.python.org}{Python programming language} using the~\href{http://mpmath.org}{\texttt{mpmath} library} for arbitrary-precision arithmetic operations.

\paragraph{Converting between base 10 and base 2 representations.} This is accomplished by~\texttt{decimalToBinary} and~\texttt{binaryToDecimal} that are both implemented below as fold expressions.  (In addition we set~$\tau \in \mathbb{N}$, as a working example, with a default value of~$\tau=8$.)

\vspace{0.3cm}

\begin{lstlisting}[language=Python, caption=]
from functools import reduce
from mpmath import mp

tau = 8

def decimalToBinary(decimalInitial, targetBinaryPrecision = tau):
    return reduce(lambda acc, _: [dyadicMap(acc[0]), acc[1] + ('0' if acc[0] < 0.5 else '1')], 
                  range(targetBinaryPrecision), 
                  [decimalInitial, ''])[1]

def binaryToDecimal(binaryInitial):
    return reduce(lambda acc, val: acc + int(val[1]) / mpmath.power(2, (val[0] + 1)), 
                  enumerate(binaryInitial), 
                  mp.mpf(0.0))

\end{lstlisting}

\vspace{0.3cm}

\noindent where~\texttt{dyadicMap}, the implementation of the dyadic transformation defined in~eq.(\ref{eq:dyadic}), can be found in the code summary paragraph of section~\ref{lab:naive}. \\

\noindent The precision of the binary strings returned by~\texttt{decimalToBinary} is kept to a small number~$\tau$ of significant bits;  as a result, this conversion can be implemented using native data types.  On the other hand, the need to go from binary to decimal representations comes from the conversion of the initial condition (constructed as a very long binary string composed of~$\sim n\tau$ significant bits encoding~$\sim n$ datapoints) to an accurate decimal value.  Keeping this level of precision when moving over to decimal representation requires the use of arbitrary-precision arithmetic operations (see section~\ref{sec::decimalPrecision}) provided by~\texttt{mpmath} methods in~\texttt{binaryToDecimal}.

\paragraph{Defining a random dataset~$\mathcal{X}$.}  In order to give concrete results to the code fragments, let us consider a dataset~$\mathcal{X}$ denoted as~\texttt{xs} made up of~50 random values in the unit interval. \\

\begin{lstlisting}[language=Python, caption=]
import numpy as np
np.random.seed(0)

numbDataPoints = 50
xs = np.random.uniform(0, 1, numbDataPoints)
print('xs = %s' % xs)

xs = [0.5488135  0.71518937 0.60276338 0.54488318 0.4236548  0.64589411
 0.43758721 0.891773   0.96366276 0.38344152 0.79172504 0.52889492
 0.56804456 0.92559664 0.07103606 0.0871293  0.0202184  0.83261985
 0.77815675 0.87001215 0.97861834 0.79915856 0.46147936 0.78052918
 0.11827443 0.63992102 0.14335329 0.94466892 0.52184832 0.41466194
 0.26455561 0.77423369 0.45615033 0.56843395 0.0187898  0.6176355
 0.61209572 0.616934   0.94374808 0.6818203  0.3595079  0.43703195
 0.6976312  0.06022547 0.66676672 0.67063787 0.21038256 0.1289263
 0.31542835 0.36371077]

\end{lstlisting}

\end{appendices}

\end{document}